\documentclass[final]{main}

\usepackage{times}
\usepackage{epsfig}
\usepackage{graphicx}
\usepackage{amsmath}
\usepackage{amssymb}
\usepackage{tabu}
\usepackage{multirow}
\usepackage{wrapfig}
\usepackage{graphicx}
\usepackage{subfigure}
\usepackage{algorithm}
\usepackage{algorithmic}
\usepackage{caption}
\usepackage{array}
\usepackage{booktabs}
\usepackage{multicol}

\usepackage[pagebackref=true,breaklinks=true,colorlinks,bookmarks=false]{hyperref}

\def\cvprPaperID{2617} 
\def\confYear{CVPR 2021}

\begin{document}

\title{Towards Compact CNNs via Collaborative Compression}

\author{
    Yuchao Li$^{1,2*}$,
    Shaohui Lin$^{3}$\thanks{Equal contribution.},
    Jianzhuang Liu$^{4}$,
    Qixiang Ye$^{5}$, 
    Mengdi Wang$^{2}$ \\
    Fei Chao$^{1}$,
    Fan Yang$^{6}$,
    Jincheng Ma$^{6}$,
    Qi Tian$^{4}$,
    Rongrong Ji$^{1,7,8}$\thanks{Corresponding author.}
    \\\\
    $^{1}$Media Analytics and Computing Laboratory, Department of Artificial Intelligence, \\ School of Informatics, Xiamen University, 
    $^{2}$Alibaba Group,
    $^{3}$East China Normal University \\
    $^{4}$Huawei Noah’s Ark Lab, 
    $^{5}$University of Chinese Academy of Sciences,
    $^{6}$Huawei TechnologiesCo.,Ltd \\
    $^{7}$Institute of Artificial Intelligence, Xiamen University,
    $^{8}$Peng Cheng Laboratory
    \\\\
    {\tt\small \{laiyin.lyc, didou.wmd\}@alibaba-inc.com,}
    {\tt\small shlin@cs.ecnu.edu.cn,}
    {\tt\small qxye@ucas.ac.cn,} \\
    {\tt\small \{liu.jianzhuang, yangfan74\}@huawei.com,}
    {\tt\small majincheng1@hisilicon.com,}
    {\tt\small \{fchao, rrji\}@xmu.edu.cn}
}

\maketitle

\begin{multicols}{2}

\begin{abstract}
  Channel pruning and tensor decomposition have received extensive attention in convolutional neural network compression.
  However, these two techniques are traditionally deployed in an isolated manner, leading to significant accuracy drop when pursuing high compression rates. 
  In this paper, we propose a Collaborative Compression (CC) scheme, which joints channel pruning and tensor decomposition to compress CNN models by simultaneously learning the model sparsity and low-rankness.
  Specifically, we first investigate the compression sensitivity of each layer in the network, and then propose a Global Compression Rate Optimization that transforms the decision problem of compression rate into an optimization problem.
  After that, we propose multi-step heuristic compression to remove redundant compression units step-by-step, which fully considers the effect of the remaining compression space (i.e., unremoved compression units).
  Our method demonstrates superior performance gains over previous ones on various datasets and backbone architectures.
  For example, we achieve 52.9\% FLOPs reduction by removing 48.4\% parameters on ResNet-50 with only a Top-1 accuracy drop of 0.56\% on ImageNet 2012.
  
\end{abstract}

\section{Introduction}

Remarkable achievements have been attained by convolutional neural networks (CNNs), such as object classification \cite{krizhevsky2012imagenet, simonyan2014very, he2016deep}, detection \cite{Ren2015Faster, Redmon2018YOLOv3} and segmentation \cite{chen2018encoder}.
However, the explosive growth of parameters and computational cost in CNN models has restricted their deployment on resource-limited devices, such as mobile or wearable devices.
To this end, extensive efforts have been made for CNN compression and acceleration, including but not limited to, parameter pruning \cite{wen2016learning, luo2017thinet, li2016pruning}, tensor decomposition \cite{lebedev2014speeding, lin2018holistic, wen2017coordinating} and quantization \cite{jacob2017quantization, zhou2016dorefa}.

\begin{figure*}[t]
\centering
\includegraphics[width=2\columnwidth]{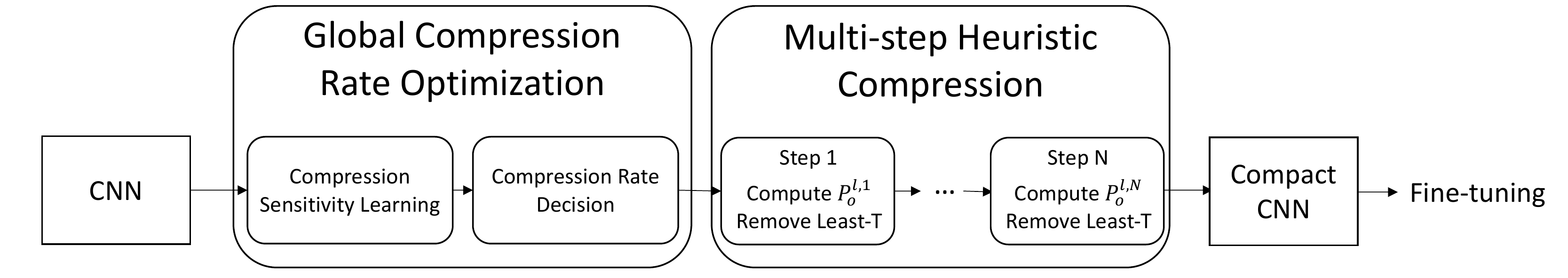}
 \caption{The framework of our method. We first compute the group values between information loss and compression rate in each layer and then obtain the compression rate of each layer by solving a global compression rate optimization problem based on them. After that, we compress each layer independently by removing the less important compression units step-by-step based on the removed units and the exploration of remaining compression space. }
  \vspace{-1.5em}
 \label{fig:framework}
\end{figure*} 

Parameter pruning and tensor decomposition are two widespread directions in CNN compression, which both aim to remove intrinsic redundancy in parameters with different removing strategies.
Parameter pruning removes correlated weight connections \cite{han2015deep, han2015learning} or structured neurons \cite{luo2017thinet, he2017channel, lin2018accelerating} based on importance measurement methods, resulting in sparse weight structures.  
In contrast, tensor decomposition approximates weights of low-rank filters based on the intrinsic low-rankness of parameters \cite{zhang2015accelerating, wen2017coordinating, lin2018holistic, kim2019efficient}.
It is thus a natural thought to combine these two compression strategies, which might lead to the significant accuracy drop when pursuing high compression rate.
For instance, Dubey \emph{et al.} \cite{dubey2018coreset} proposed to compress the weights by sequentially employing pruning and tensor decomposition, which assumes that they are complementary without any mutual influence.
However, as demonstrated in previous work \cite{yu2017compressing}, although the pruning and decomposition explore different redundancy in parameters, they are not completely orthogonal.
Thus, the above method \cite{dubey2018coreset} does not exploit the complementary nature of pruning and decomposition, which is sub-optimal as exploring only within each sub-task (\emph{i.e.,} each compression method), not from the global compression scope.

To leverage the benefits of both compression operations, training-aware methods \cite{yu2017compressing, ma2019unified, li2020group} use two regularizations to separately handle the sparsity on channel pruning and the low-rankness on tensor decomposition, which are jointly minimized the regularization loss during training.
They show that simultaneously handle sparsity and low-rankness in weights can explore the richer structure information of parameters.
However, these methods are difficult to control the compression rate, which needs to adjust hyper-parameters by trial-and-error to achieve the trade-off between compression rate and model accuracy.

In this paper, we propose a novel unified compression framework, named \emph{collaborative compression} (CC), to simultaneously deal with the sparsity and low-rankness in weights, with an essential innovation in automatic compression rate control.
Our method is a post-training compression algorithm that simultaneously explores the sparsity and low-rankness in pre-trained networks, which is easier to apply than the training-aware methods.
As shown in Fig.~\ref{fig:framework}, towards a fast and practical compression, we first determine the compression rate of each layer by \emph{global compression rate optimization} and then compress each layer independently by \emph{multi-step heuristic compression}.
It avoids determining the compression rate and the compression strategy (\emph{i.e.,} which compression units should be removed) simultaneously.
In particular, the global compression rate optimization analyzes the compression sensitivity of each layer by constructing the relationship between the proposed information loss and the compression rate.
We find that this relationship can be fitted well by an exponential function, and the compression sensitivity can be viewed as the first-order derivative of this exponential function.
Based on the exponential model, we construct an optimization process to search the best compression rate of each layer. 
After determining the compression rate, we compress each layer synchronously and independently.
Considering that removing units will affect the importance of the remaining compression units, we propose a multi-step heuristic compression method, which removes less important units step-by-step.
At each step, the importance calculation of each unit fully considers the effect of removed compression units and the remaining compression space.

In experiments, we demonstrate the effectiveness of the proposed CC framework using four widely-used networks (VGGNet, GoogleNet, ResNet and DenseNet) on two datasets (CIFAR-10 and ImageNet 2012).
Compared to the state-of-the-art methods, CC achieves superior performance.
For example, we obtain 52.9\% FLOPs reduction by removing 48.4\% parameters, with only a Top-1 accuracy drop of 0.56\% on ResNet-50.
Meanwhile, the compressed MobileNet-V2 obtained by our method achieves performance gains over the state-of-the-art pruning methods based on AutoML.

\begin{figure*}[t]
 \subfigure[Resnet56: B2-U8-C2]{
    \includegraphics[width=0.49\columnwidth]{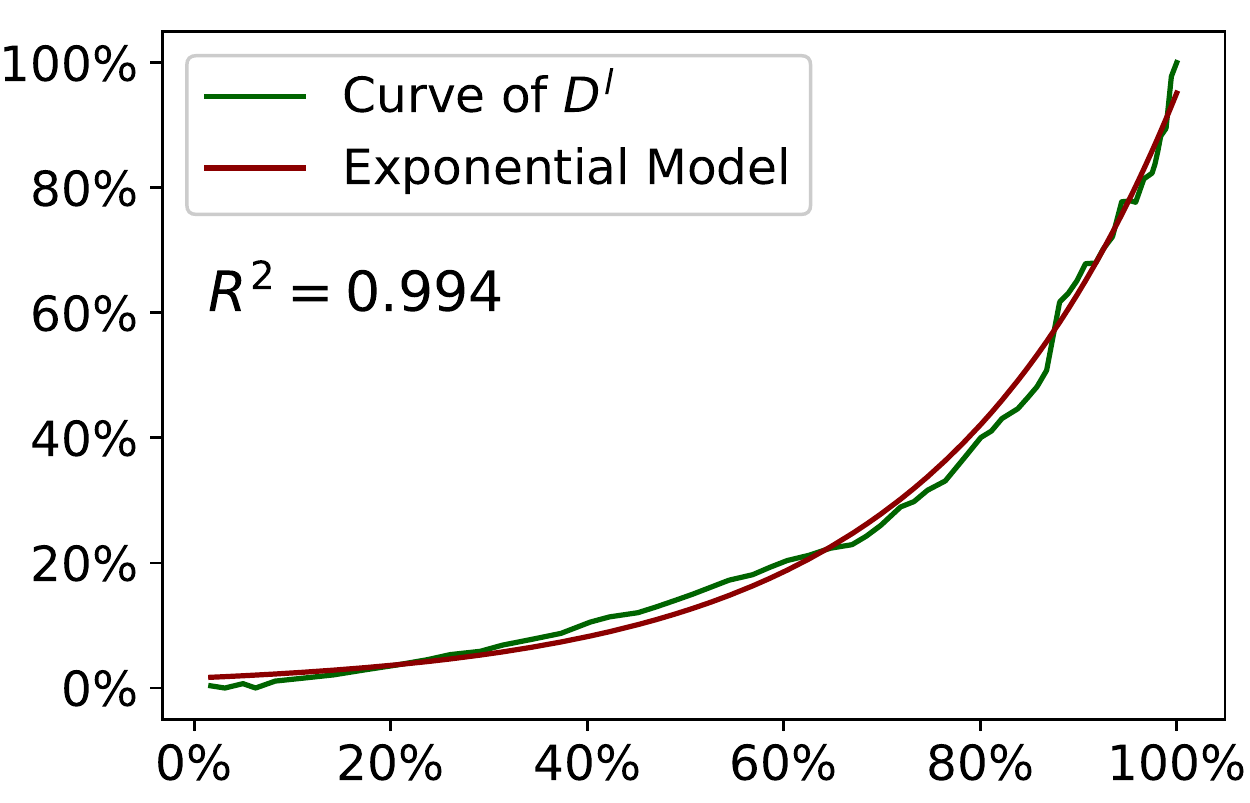}
    }
  \hspace{-1em}
 \subfigure[DenseNet40: Block1-Conv5]{
    \includegraphics[width=0.49\columnwidth]{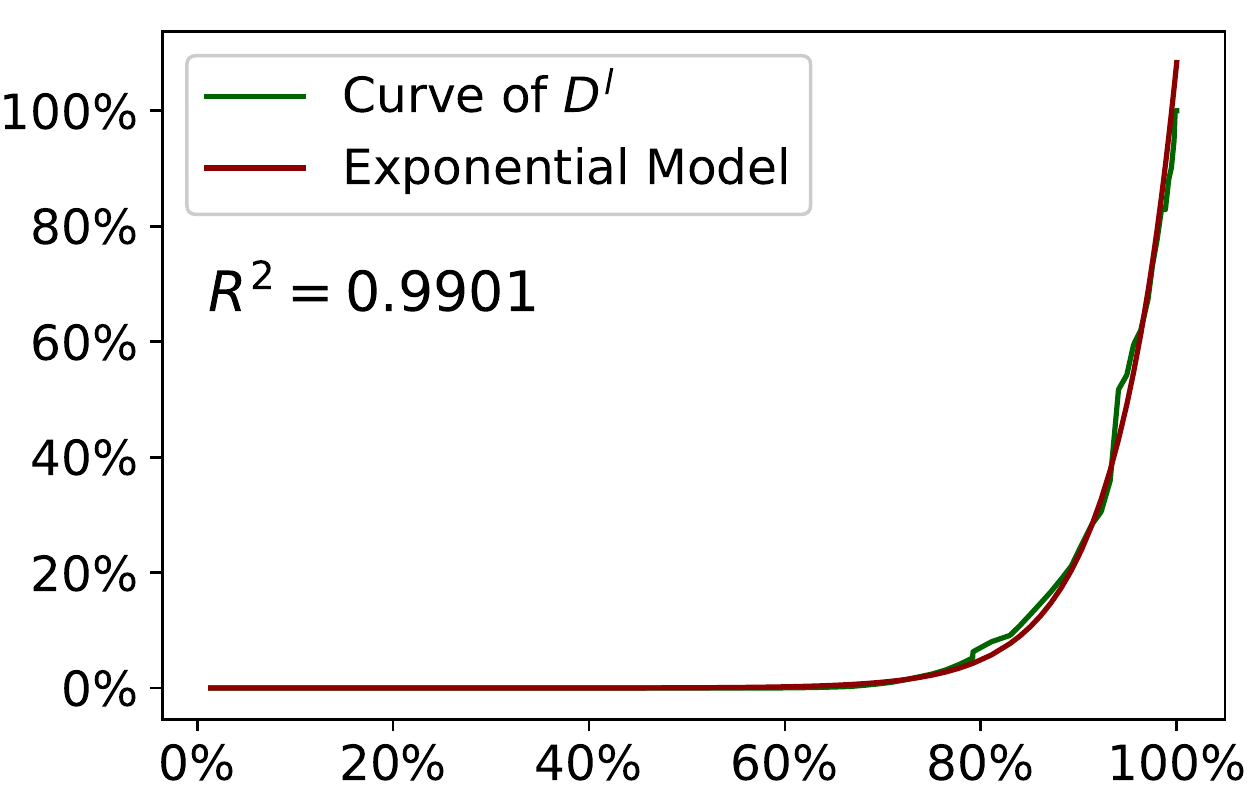}
  } 
  \hspace{-1em}
  \subfigure[VGGNet: Conv3]{
    \includegraphics[width=0.49\columnwidth]{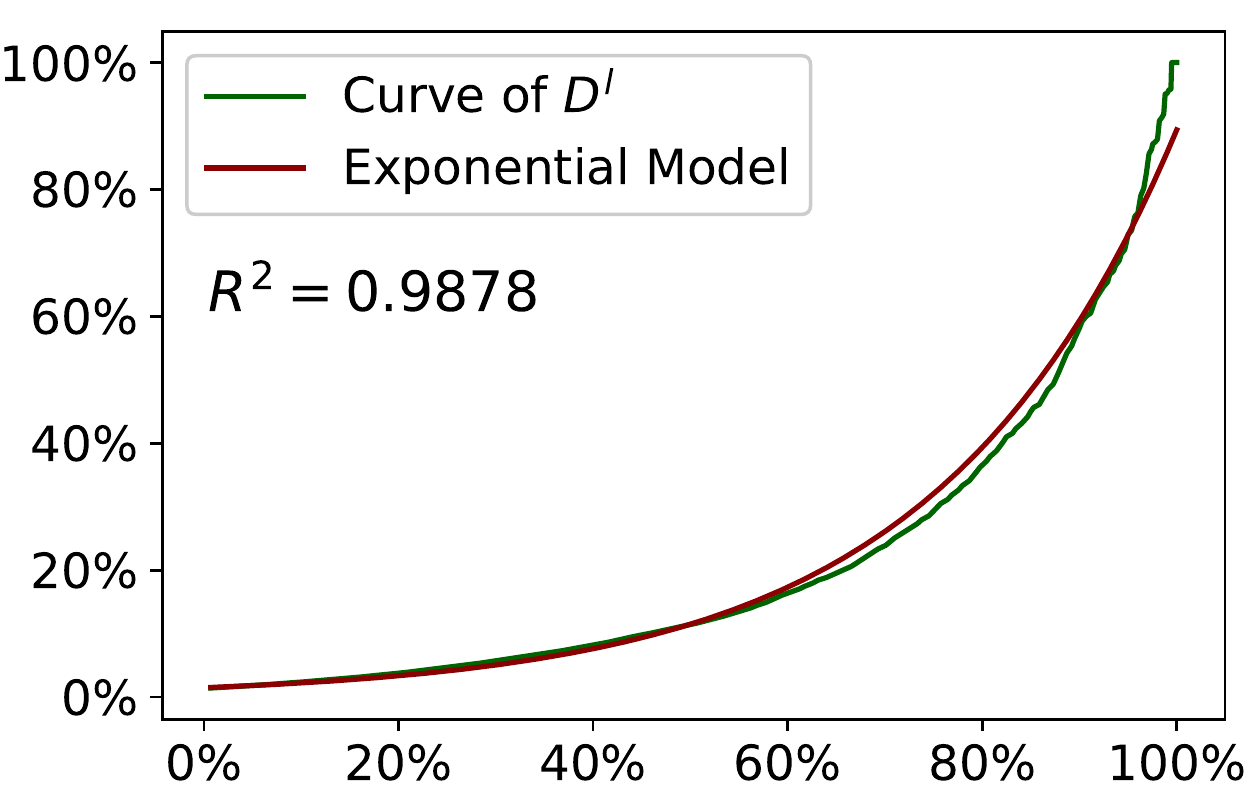}
  } 
  \hspace{-1em}
  \subfigure[GoogLeNet: A3-B4-C1]{
    \includegraphics[width=0.49\columnwidth]{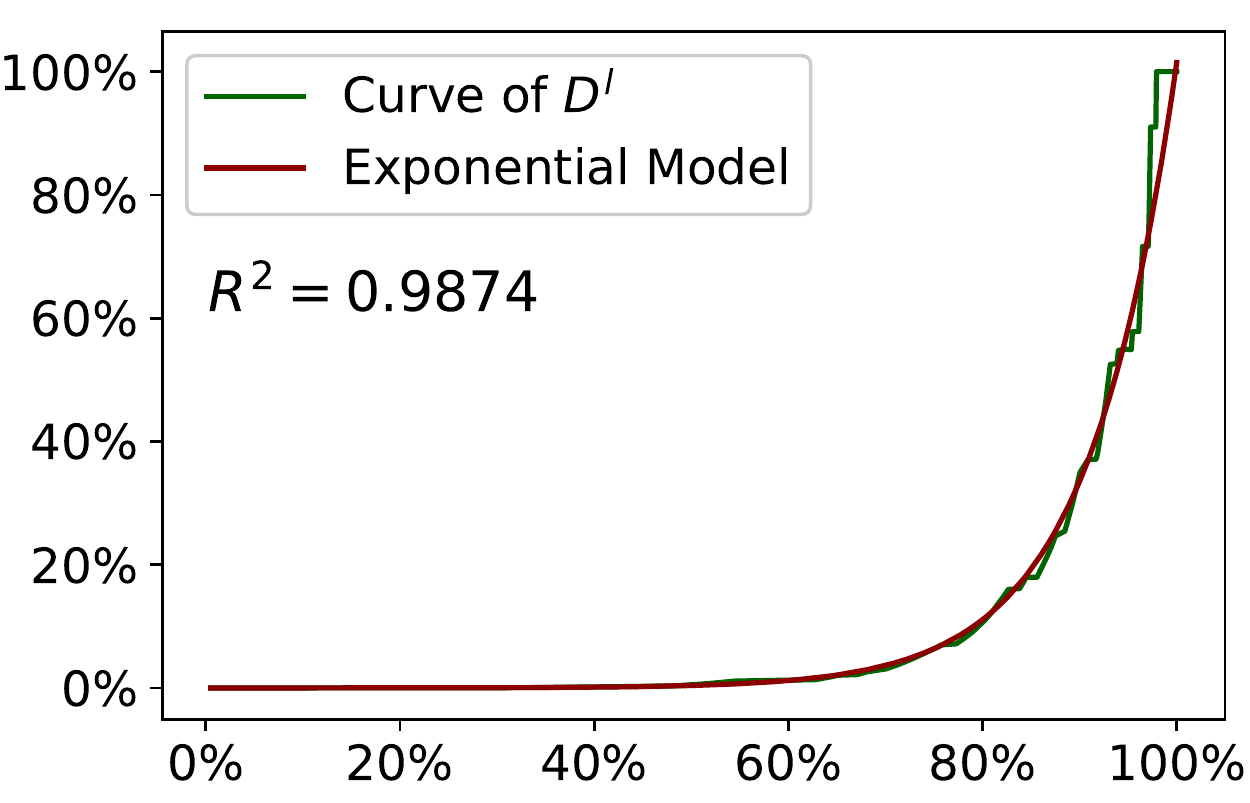}
  } 
  \hspace{-1em}
  \subfigure[Resnet56:B3-U8-C2]{
  \includegraphics[width=0.49\columnwidth]{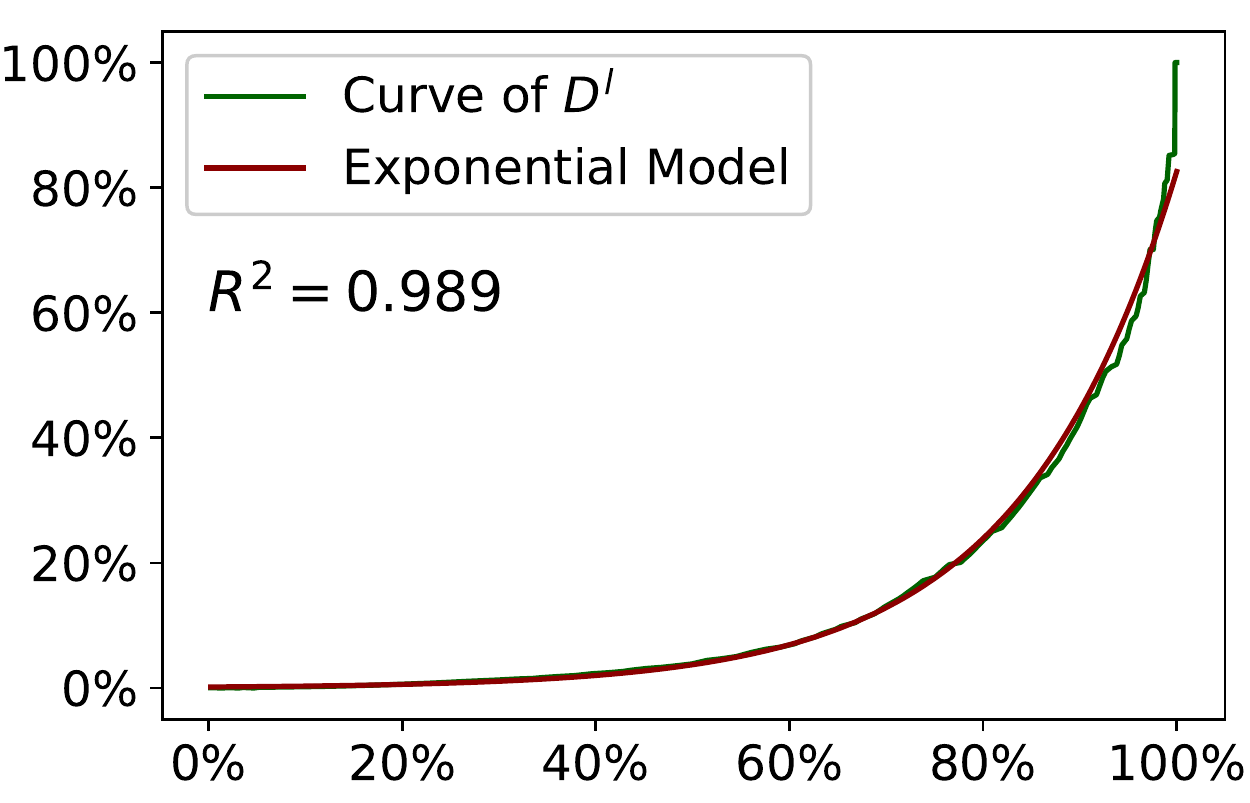}
  }
  \subfigure[DenseNet40:Block2-Conv5]{
    \includegraphics[width=0.49\columnwidth]{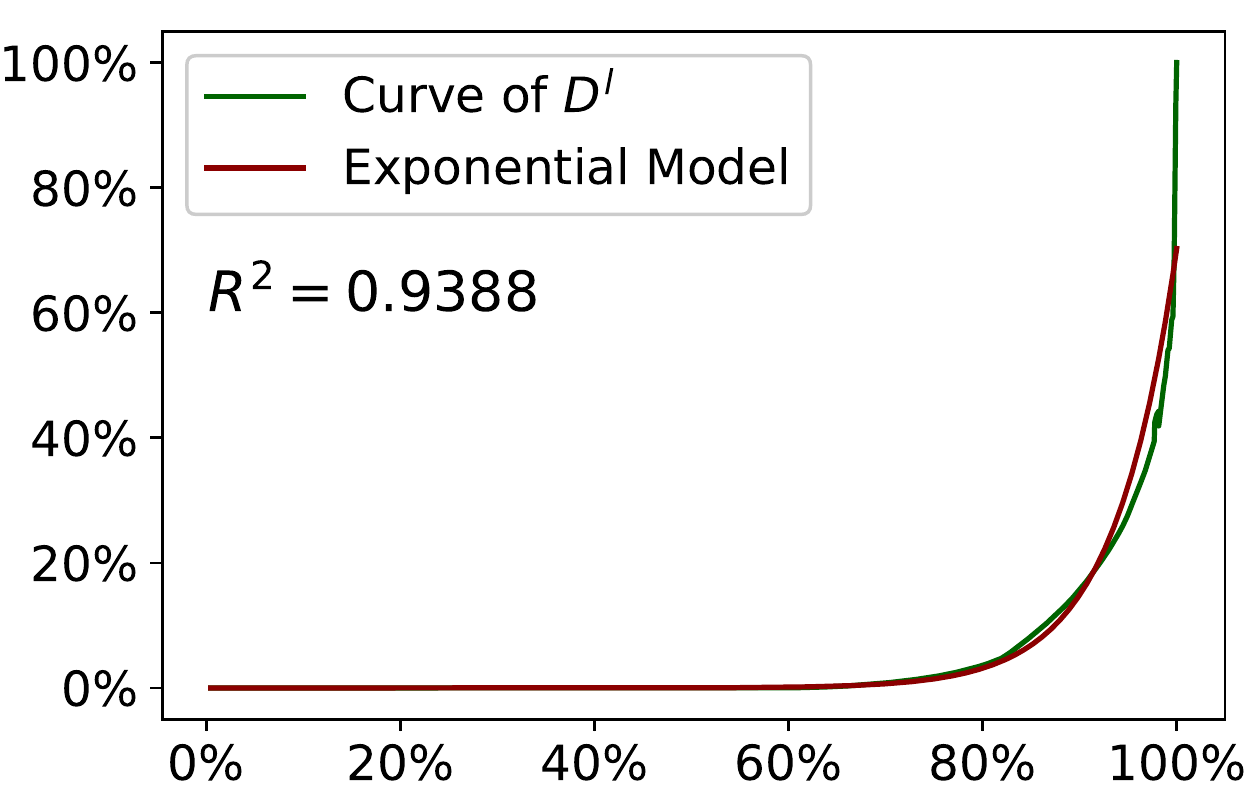}
  } 
  \subfigure[VGGNet:Conv4]{
    \includegraphics[width=0.49\columnwidth]{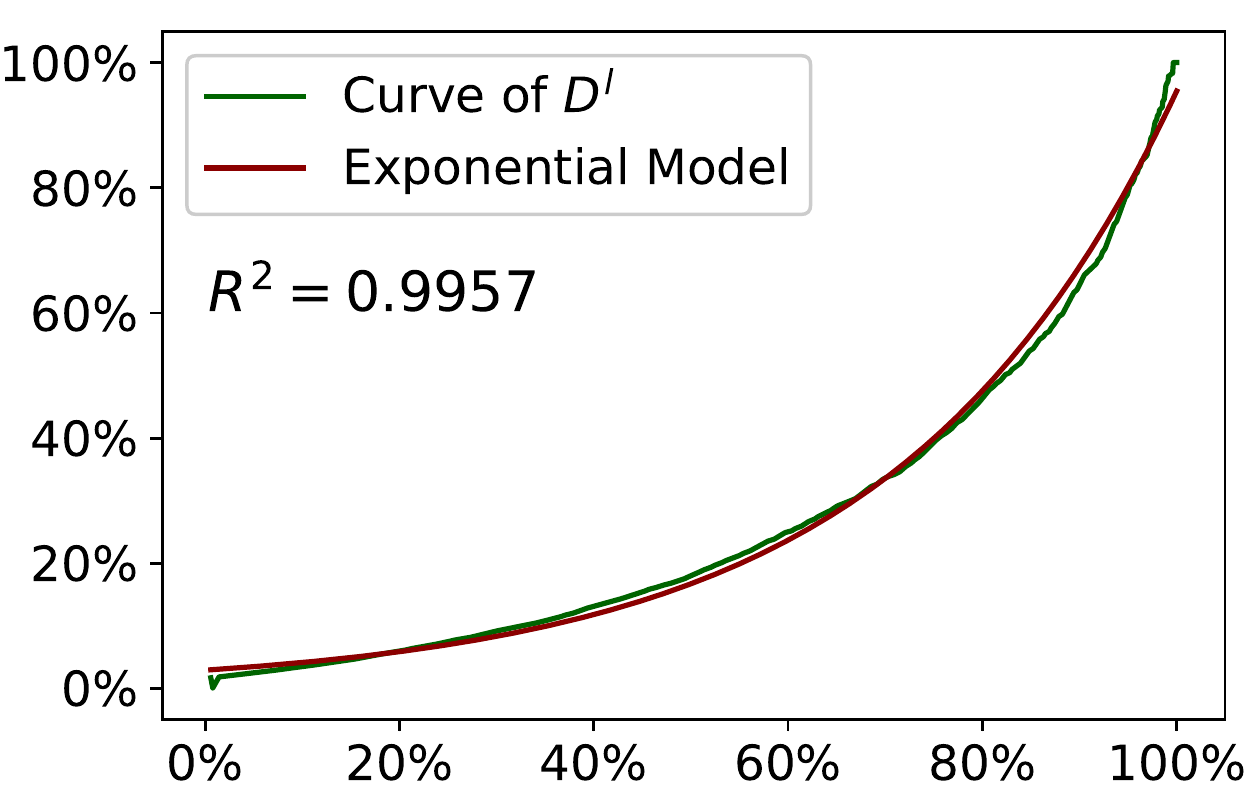}
  } 
  \subfigure[GoogLeNet:A4-B4-C1]{
    \includegraphics[width=0.49\columnwidth]{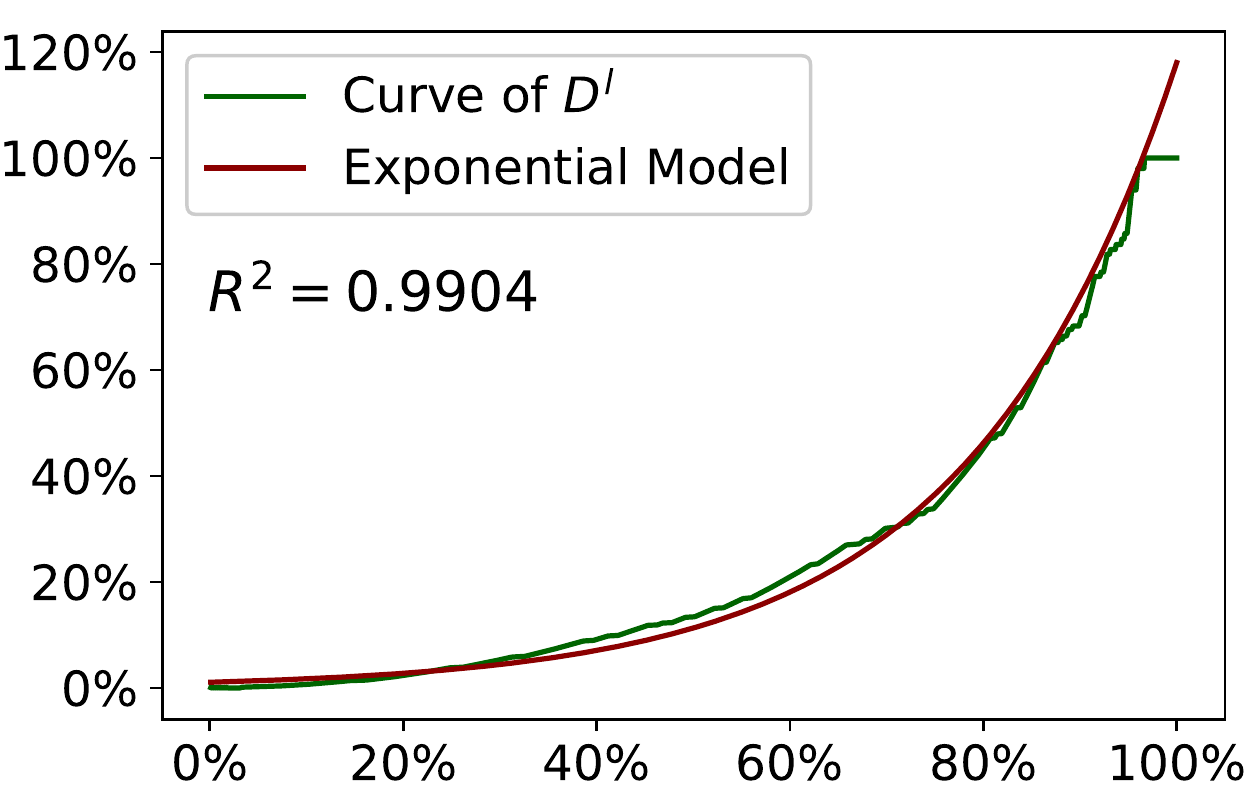}
  } 
  \vspace{-1.2em}
 \caption{Compression sensitivity analysis of different convolutional layers in different networks on CIFAR-10. For each subfigure, the x-axis represents the compression rate and the y-axis is the information loss based on Eq.~\ref{eq:importance}, both of which are normalized to $[0, 1]$. $R^2$ represents the coefficient of determination between $\mathcal{D}^l$ and the learned exponential model, which measures the result of model fitting. $R^2$ closer to 1 is better.}
 \vspace{-1.2em}
 \label{fig:csc}
  \end{figure*}  

\section{Related Work}

Pruning can be categorized into either unstructured or structured pruning.
Unstructured pruning \cite{han2015deep, han2015learning} aims to remove unimportant weights independently, while structured pruning removes structured parts (\emph{e.g.,} filters, channels or layers) that are well supported by various off-the-shelf deep learning libraries.
Structured pruning, especially filter pruning, removes redundant filters by different importance measurements, such as $\ell_1$-norm on filters \cite{li2016pruning, he2018amc},  sparsity of output feature maps \cite{hu2016network}, sparsity regularization \cite{wen2016learning, ding2019centripetal, lin2019toward} and loss drop \emph{w.r.t.} filter removal \cite{molchanov2019importance, luo2017thinet}.
Similar to filter pruning, channel pruning \cite{liu2017learning, huang2018condensenet} targets at removing input channels, which can avoid the dimensional mismatch in various multi-branch networks, \emph{e.g.,} ResNets and DenseNets.
Apart from this, layer pruning has been proposed for data-dependent inference \cite{veit2018convolutional, veit2016residual} or static block removal \cite{lin2019towards}.

Tensor decomposition \cite{wen2017coordinating, li2018constrained, lin2018holistic, kim2019efficient} aims to reduce the memory and computation cost by decomposing the convolutional filters into a sequence of tensors with fewer parameters.
Unlike pruning, it explores the low-rank structure of the original weights, which keeps the dimension of convolutional outputs unchanged.
Early works directly introduce different tensor decomposition methods on the original weights, such as SVD-decomposition \cite{zhang2015accelerating}, Tucker-decomposition \cite{kim2015compression} and CP-decomposition \cite{lebedev2014speeding}.
Later on, Lin \emph{et al.} \cite{lin2018holistic} proposed a novel closed-form low-rank decomposition for fast decomposition.
Different from direct decomposition, Wen \emph{et al.} \cite{wen2017coordinating} proposed a force regularization to coordinate and learn more weight information into a low-rank space.
Compared to these single compression operation-based methods (\emph{i.e., only using one compression operation}), we propose a collaborative compression method, which focuses on simultaneously reducing the sparsity and low-rankness in the weights and fully considers the mutual influence between them.

To combine multiple compression operations, previous studies \cite{han2015deep, dubey2018coreset} proposed multiple compression stages, which independently adopt one compression operation at each step and ignore the mutual influence between different compression operations.
For instance, Dubey \emph{et al.} \cite{dubey2018coreset} first adopted filter pruning to compress the weights and then decompose them based on a Coreset-based decomposition method.
Moreover, some training-aware compression methods \cite{yu2017compressing, tung2018clip, li2020group} compress networks during training by using regularization.
Li \emph{et al.} \cite{li2020group} first introduced a sparsity-inducing matrix followed at each weight and then added group sparsity constraints on them during training.
However, these training-aware methods are difficult to search a good trade-off between compression rate and accuracy under the target compression rate.
In contrast, our collaborative compression employs the global compression rate optimization method to easily obtain the compression rate of each layer under the target compression rate.

\vspace{-1em}
\section{Proposed Method}

\subsection{Preliminaries}

Generally, the $l$-th convolutional layer in CNNs transforms an input tensor $\mathcal{X}^l \in \mathbb{R}^{c^l\times h^l_{in}\times w^l_{in}}$ into an output tensor $\mathcal{Y}^l \in \mathbb{R}^{n^l\times h^l_{out}\times w^l_{out}}$ by using a weight tensor $\mathcal{W}^l \in \mathbb{R}^{n^l \times c^l \times k^l \times k^l}$, where $c^l$ and $n^l$ denote the number of input channels and filters (output channels), respectively, and $k^l\times k^l$ is the spatial size of the filters.
The convolution operation is represented as:
%
\begin{equation}
\small
  \label{method:eq_1}
\mathcal{Y}^l = \mathcal{W}^l \otimes \mathcal{X}^l,
\end{equation}
%
where $\otimes$ denotes the convolution operation.
The biases are omitted for simplicity.
Both channel pruning and tensor decomposition seek a compact approximated representation $\overline{\mathcal{W}}^l \in \mathbb{R}^{n^l \times c^l \times k^l \times k^l}$ to replace $\mathcal{W}^l$.
Thus, the compression process is regarded as a function $\overline{\mathcal{W}}^l = f(\mathcal{W}^l, o)$, where $o$ represents a single unit index, whose corresponding unit should be removed.

\textbf{Channel Pruning.} 
We introduce input channel pruning, where each input channel is regarded as a compression unit.
Therefore, the compression function is defined as:
\begin{equation}
\small
\label{eq:channel_pruning}
\overline{\mathcal{W}}^l_{:, i, :, :} =
\left\{
\begin{aligned}
&0 , &i = o, \\
&\mathcal{W}^l_{:, i, :, :} , &i \neq o.
\end{aligned}
\right.
\end{equation}

\textbf{Tensor Decomposition.}
We decompose the network's weights based on the Singular Value Decomposition (SVD), where the original $\mathcal{W}^l$ is first mapped from a tensor to a matrix $M^l \in \mathbb{R}^{n^l \times (c^l k^l k^l)}$.
After that, let $M^l = U^l \Sigma^l V^{l^{\top}}$ be the SVD of $M^l$, where $U^l\in\mathbb{R}^{n^l\times n^l}$ and $V^l\in\mathbb{R}^{c^lk^lk^l\times c^lk^lk^l}$ are two orthogonal matrices.
The diagonal elements in $\Sigma^l\in\mathbb{R}^{n^l\times (c^lk^lk^l)}$ are singular values of $M^l$, whose number is $r^l=min(n^l, c^lk^lk^l)$.
Finally, the weight matrix $M^l$ is decomposed into two light matrices $M^l_1 = \sqrt{\Sigma^l} V^{l^{\top}}$ and $M^l_2 = U^l \sqrt{\Sigma^l}$ when the number of non-zero singular values $\bar{r}^l$ is much smaller than $r^l$.  
For implementation, $M_1^l$ and $M_2^l$ are reshaped back into $\mathcal{W}_1^l \in \mathbb{R}^{\overline{r}^l \times c^l \times k^l \times k^l}$ and $\mathcal{W}_2^l \in \mathbb{R}^{n^l \times \overline{r}^l \times 1 \times 1}$, respectively.
Thus, two light convolutions are computed to approximate the original convolution by $\mathcal{W}_2^l\otimes\mathcal{W}_1^l\otimes\mathcal{X}^l$.
The compression rate is controlled by the number of non-zero singular values $\overline{r}^l$ in $\Sigma^l$.
Therefore, the compression unit index is the index of the singular values, and the compression function is defined as:
\begin{equation}
\small
\label{eq:tensor_decomposition}
f(\mathcal{W}^l, o) = \overline{\mathcal{W}}^l = \phi(U^l \overline{\Sigma}^l V^{l^{\top}}),\   
\overline{\Sigma}^l_{i,i} =
\left\{
\begin{aligned}
&0 , &i = o, \\
&\Sigma^l_{i,i} , &i \neq o,
\end{aligned}
\right.
\end{equation}
where $\phi : \mathbb{R}^{n^l \times (c^l k^l k^l)} \mapsto \mathbb{R}^{n^l \times c^l \times k^l \times k^l}$ is a mapping function (\emph{a.k.a.} reshaping operator).

In summary, $W^l$ has $c^l + r^l$ candidate compression units.
We can obtain compressed weights $\widetilde{\mathcal{W}}^l$ if removing $t_1$ input channels and $t_2$ singular values:
\begin{equation}
\small
\label{eq:compressed_weight}
\widetilde{\mathcal{W}}^l = \left\{
\begin{array}{ll}
\widetilde{\mathcal{W}}_1^l \in \mathcal{R}^{(r^l-t_2) \times (c^l-t_1) \times k^l \times k^l}, & \multirow{2}*{$t_2\neq 0$} \\  \widetilde{\mathcal{W}}_2^l \in \mathcal{R}^{n^l \times (r^l-t_2) \times 1 \times 1}, \\
\specialrule{0em}{1ex}{1ex}
\widetilde{\mathcal{W}}^l \in \mathcal{R}^{n^l \times (c^l-t_1) \times k^l \times k^l}, &t_2 = 0,
\end{array}
\right.
\end{equation}
where $t_2 = 0$ represents that we only adopt channel pruning to compress the convolutional weights.
In addition, the compression rate at the $l$-th layer is defined as:
\begin{equation}
\small
\label{eq:compression_ratio}
R^l = 
\left\{
\begin{aligned}
&1 - \frac{(r^l - t_2)*[(c^l-t_1)*k^l*k^l + n^l]}{n^l*c^l*k^l*k^l} , &t_2 \neq 0, \\
&\frac{t_1}{c^l} , &t_2 = 0.
\end{aligned}
\right.
\end{equation}

\subsection{Global Compression Rate Optimization}

As presented in \cite{li2016pruning, lin2019toward}, since the compression sensitivity of each layer is different, the compression rate of the sensitive layers should be set to a small value while the compression rate of the insusceptible layers should be set to a large one.
To obtain the compression sensitivity of the layers, the method \cite{li2016pruning} first removes a certain ratio of less important compression units in the weights and then evaluates the information loss.
Such process needs to be evaluated iteratively with different compression rates to obtain the curve between compression rate and information loss, which is used to visualize the compression sensitivity.
It inevitably makes the method labor-intensive and costs plenty of time with human analysis to determine the best compression rate.
In this paper, we first fast evaluate the compression sensitivity of each layer and then search the best compression rate by solving a simple optimization problem.

Firstly, the information loss at the $l$-th layer is defined to indicate how much the network loss increases when removing the compression units in this layer.
Inspired by \cite{molchanov2016pruning, molchanov2019importance}, we measure the information loss at the $l$-th layer by adopting the first-order Taylor-based approximation of the network loss on the compressed weights $\mathcal{W}^l$:

\begin{algorithm}[H]
\small
 \caption{\small{Compression sensitivity learning algorithm}}
 \label{alg:1}
 \begin{algorithmic}[1]
   \REQUIRE{The pre-trained weights $\mathcal{W}^l$ at the $l$-th layer and the corresponding average gradients $\mathcal{G}^l$.}
   \ENSURE{An exponential model $I^l = a^l e^{b^l R^l}$.}
   
   \STATE Initialize the set of compression unit indices $\mathcal{U}^l$, whose corresponding unit number is $c^l+r^l$. \\
   \FOR {each compression unit index $o$ in $\mathcal{U}^l$}
        \STATE $I^l_{o} = || \mathcal{G}^l * (f(\mathcal{W}^l, o) - \mathcal{W}^l)||^2_2$.\\
  \ENDFOR
    \STATE Sort $\mathcal{U}^l$ based on $I^l$ ascendingly. \\
    \STATE Initialize $\overline{\mathcal{W}}^l = \mathcal{W}^l$, $\mathcal{D}^l=\emptyset$. \\
    \FOR {each compression unit index $o$ in the sorted $\mathcal{U}^l$}
        \STATE $\overline{\mathcal{W}}^{l} = f(\overline{\mathcal{W}}^{l}, o)$. \\
        \STATE $I^l = \frac{|| \mathcal{G}^l * (\overline{\mathcal{W}}^l - \mathcal{W}^l)||^2_2}{|| \mathcal{G}^l * \mathcal{W}^l||^2_2}$. \\
    \STATE Compute $R^l$ via Eq.~\ref{eq:compression_ratio}. \\
    \STATE Add $(R^l, I^l)$ into $\mathcal{D}^l$. \\
  \ENDFOR
    \STATE Using the least squares method to fit $\mathcal{D}^l$ by the exponential model $I^l = a^l e^{b^l R^l}$. \\
 \end{algorithmic}
\end{algorithm} 

\vspace{-2em}
\begin{algorithm}[H]
 \caption{Compression rate decision algorithm}
 \label{alg:2}
 \begin{algorithmic}[1]
   \REQUIRE{The evaluation model of each layer $\{I=a^1e^{b^1R},...,I=a^Le^{b^LR}\}$, the FLOPs of each layer $\{F^1,...,F^L\}$, the FLOPs of the entire network $F$, the target compression rate $C$, and learning rate $\eta$. }
   \ENSURE{The target compression rate of each layer $\{R^1,...,R^L\}$.}
   
   \STATE Initialize $\overline{I}' = 0.1$. \\
   \WHILE{$(\sum\limits_{i=1}^{L} \frac{F^i}{b^i} log(\frac{\overline{y}'}{a^i b^i}) - C F)^2 > 10^4$}
    \STATE $g = 2(\sum\limits_{i=1}^{L} \frac{F^i}{b^i} log(\frac{\overline{I}'}{a^i b^i}) - C F)(\sum\limits_{i=1}^{L}\frac{F^i}{b^i\overline{I}'})$. \\
    \STATE $\overline{I}' = \overline{I}' - \eta g$.
   \ENDWHILE
   \RETURN $\{R^i = \frac{1}{b^i} log(\frac{\overline{I}'}{a^i b^i})\}_{i=1}^{L}$.
 \end{algorithmic}
\end{algorithm}
\vspace{-1em}

\begin{equation}
\begin{aligned}
\small
\label{eq:importance}
  I^l = [L(\overline{\mathcal{W}}^l)-L(\mathcal{W}^l)]^2 
  &\approx \sum_{i \in Q^l}(\frac{\partial \mathcal{L}}{\partial \mathcal{W}^l_{i}} * (\overline{\mathcal{W}}^l_{i}-\mathcal{W}^l_{i}))^2 \\
  &= \mathrm{S}[(\mathcal{G}^l * (\overline{\mathcal{W}}^l -\mathcal{W}^l))^2],
\end{aligned}
\end{equation}
where $\mathcal{G}^l \in \mathbb{R}^{n^l \times c^l \times k^l \times k^l}$ denotes the gradient computed by the average gradient of the loss \emph{w.r.t.} the pre-trained weights $W^l$ over the entire dataset, $Q^l$ is all element indices in $\mathcal{W}^l$, and $\mathrm{S}[\cdot]$ denotes the sum of all elements in the tensor.

After that, we propose a greedy algorithm shown in Alg.~\ref{alg:1} to fast evaluate the information loss under different compression rates, which is significantly fast to measure compression sensitivity of each layer.
We first compute the importance of each compression unit using Eq.~\ref{eq:importance}.
Then, to obtain the information loss under different compression rates (\emph{i.e.,} a point set $\mathcal{D}^l$), we remove candidate compression units, based on their importance progressively, and compute the corresponding information loss $I^l$ and compression rate $R^l$ after each unit removed.
As shown in Fig.~\ref{fig:csc}, the green lines are from the point set $\mathcal{D}^l$ where $R^l$ and $I^l$ are normalized to $[0, 1]$.
The results show that the compression sensitivity of different layers are possibly variable, such as Fig.~\ref{fig:csc}(a) vs. Fig.~\ref{fig:csc}(e) in ResNet-56.
Through calculating all the layers across various networks, we empirically find that the curve of $\mathcal{D}^l$ can be estimated by an exponential function, $I^l = a^l e^{b^l R^l}$.
Therefore, we fit the function to learn the parameters $a^l$ and $b^l$ by using the least-squares method.
Also in Fig.~\ref{fig:csc}, we find that the curve of $\mathcal{D}^l$ is well approximated by our exponential function with extremely small reconstruction error.
More results are shown in the supplementary materials. 

The compression sensitivity at the $l$-th layer can be defined by the gradient of the exponential function \emph{w.r.t} the compression rate $R^l$ as $\frac{\partial I^l}{\partial R^l} = a^lb^l e^{b^l R^l}$.
We construct the following optimization problem to decide the compression rate of each layer when given the whole network compression rate $C$:
\vspace{-1em}
\begin{equation}
\small
\label{eq:global_optim}
  \begin{aligned}
& \mathop{\min}_{\mathcal{R}} (\sum_{i=1}^L F^i \cdot R^i - C\cdot F)^2, \\ 
& s.t.\  \ a^ib^i e^{b^i R^i} = a^jb^j e^{b^j R^j},\ \forall i, j \in \{1,2...,L\},
  \end{aligned}
\end{equation}
where $\mathcal{R} = \{R^1, R^2,..., R^L\}$ denotes the set of compression rates for the layers, $F^l$ and $F$ represents the FLOPs of the $l$-th layer and the entire network, respectively.
Under the same compression rate, the compression sensitivity $\frac{\partial I}{\partial R}$ of the sensitive layers is higher than that of the insusceptible layers.
Correspondingly, the sensitive layers have a smaller compression rate than the insusceptible layers under the same compression sensitivity. 
Therefore, in the constraint term, the compression sensitivity of all layers are set to the same to make the compression rate of the sensitive layers smaller than that of the insusceptible layers.
To simplify the optimization, we formulate the following equation as:
\begin{equation}
\small
\label{eq:cs_to_cr}
  \overline{I}' = \frac{\partial I^l}{\partial R^l} = a^l b^l e^{b^l R^l} \Rightarrow R^l = \frac{1}{b^l} log(\frac{\overline{I}'}{a^l b^l}).
\end{equation}
Therefore, we can transform the optimized variables from $\mathcal{R}$ to $\overline{I}'$ according to the constraint term in Eq.~\ref{eq:global_optim}.
Thus, Eq.~\ref{eq:global_optim} is rewritten as:
\begin{equation}
\small
  \mathop{\min}_{\overline{I}'} (\sum_{i=1}^{L} \frac{F^i}{b^i} log(\frac{\overline{I}'}{a^i b^i}) - C \cdot F)^2.
\end{equation}
This optimization problem can be directly solved by a gradient descent algorithm.
The detailed optimization process is presented in Alg.~\ref{alg:2}.
After obtaining the optimal sensitivity (\emph{i.e.,} final result of $\overline{I}'$), we can also obtain the compression rate of each layer via Eq.~\ref{eq:cs_to_cr}.

\begin{table*}
  \footnotesize
\begin{center}
\begin{tabular}{|p{2cm}<{\centering}|p{3cm}<{\centering}|p{2cm}<{\centering}|p{2cm}<{\centering}|p{2cm}<{\centering}|}
\hline
Model & Method & FLOPs(PR) & \#Param.(PR) & Top-1 Acc\% \\
\hline
\multirow{11}*{ResNet-56} & Baseline & 125M & 0.85M & 93.33 \\

 & L1\cite{li2016pruning} & 112M(10.4\%) & 0.77M(9.4\%) & 93.10 \\

 & HRank\cite{lin2020hrank} & 89M(28.8\%) & 0.71M(16.5\%) & 93.52 \\

 & Nisp\cite{yu2018nisp} & 81M(35.2\%) & 0.49M(42.4\%) & 93.01 \\

 & GAL\cite{lin2019towards} & 78M(37.6\%) & 0.75M(11.8\%) & 92.98 \\

 & \textbf{CC ($C = 0.4$)} & \textbf{72M(42.4\%)} & \textbf{0.54M(36.5\%)} & \textbf{93.87} \\
 
 & ENC\cite{kim2019efficient} & 63M(49.6\%) & - & 93.09 \\
 
 & CP\cite{he2017channel} & 62M(50.4\%) & - & 91.80 \\

 & \textbf{CC ($C = 0.5$)} & \textbf{60M(52.0\%)} & \textbf{0.44M(48.2\%)} & \textbf{93.64} \\

 & FPGM\cite{he2019filter} & 59M(52.8\%) & - & 93.49 \\

 & C-SGD\cite{ding2019centripetal} & 49M(60.8\%) & - & 93.44 \\
\hline

\multirow{8}*{DenseNet-40} & Baseline & 283M & 1.04M & 94.81 \\

 & GAL\cite{lin2019towards} & 183M(35.3\%) & 0.67M(35.6\%) & 94.61 \\

 & HRank\cite{lin2020hrank} & 167M(41.0\%) & 0.66M(36.5\%) & 94.24 \\

 & \textbf{CC ($C = 0.4$)} & \textbf{150M(47.0\%)} & \textbf{0.50M(51.9\%)} & \textbf{94.67} \\

 & slimming\cite{liu2017learning} & 120M(57.6\%) & 0.35M(66.3\%) & 94.35 \\

 & C-SGD\cite{ding2019centripetal} & 113M(60.1\%) & - & 94.37 \\

 & \textbf{CC ($C = 0.6$)} & \textbf{112M(60.4\%)} & \textbf{0.37M(64.4\%)} & \textbf{94.40} \\

 & GAL\cite{lin2019towards} & 78M(72.4\%) & 0.75M(27.9\%) & 92.98 \\
\hline

\multirow{7}*{VGGNet} & Baseline & 313M & 14.72M & 93.70 \\

 & L1\cite{li2016pruning} & 206M(34.2\%) & 5.40M(63.3\%) & 93.60 \\

 & GAL\cite{lin2019towards} & 189M(39.6\%) & 3.36M(77.2\%) & 93.77 \\
  
 & AOFP\cite{ding2019approximated} & 186M(40.6\%) & - & 94.03 \\

 & \textbf{CC ($C = 0.5$)} & \textbf{154M(50.8\%)} & \textbf{5.02M(65.9\%)} & \textbf{94.15} \\

 & HRank\cite{lin2020hrank} & 145M(53.7\%) & 2.51M(82.9\%) & 93.42 \\

 & \textbf{CC ($C = 0.6$)} & \textbf{123M(60.7\%)} & \textbf{4.02M(72.7\%)} & \textbf{94.09} \\
\hline

\multirow{6}*{GoogLeNet} & Baseline & 1.52B & 6.15M & 95.05 \\

 & L1\cite{li2016pruning} & 1.02B(32.9\%) & 3.51M(42.9\%) & 94.54 \\

 & GAL\cite{lin2019towards} & 0.94B(38.2\%) & 3.12M(49.3\%) & 93.93 \\

 & \textbf{CC ($C = 0.5$)} & \textbf{0.76B(50.0\%)} & \textbf{2.83M(54.0\%)} & \textbf{95.18} \\

 & HRank\cite{lin2020hrank} & 0.69M(54.6\%) & 2.74M(55.4\%) & 94.53 \\

 & \textbf{CC ($C = 0.6$)} & \textbf{0.61B(59.9\%)} & \textbf{2.26M(63.3\%)} & \textbf{94.88} \\
\hline
\end{tabular}
\end{center}
\vspace{-2em}
\caption{Comparison with single compression operation-based methods on CIFAR-10. In this table and all following tables and figures, M/B means million/billion, and PR denotes pruning rate.}
\vspace{-2em}
\label{tab:cifar10}
\end{table*}

\subsection{Multi-Step Heuristic Compression}

After determining the compression rate, we propose a heuristic method to compress each layer's weights.
As discussed above, the importance of units will change due to the removal of other units because of the mutual influence between different compression operations.
Therefore, the previous importance metric \cite{he2019filter, lin2020hrank, molchanov2019importance} is no longer effective by only taking fixed weights into consideration and only computing the importance metric once.
Inspired by the value function in the Markov Decision Process, we propose a novel importance metric, in which the importance of the $o$-th compression unit in the $l$-th layer at the $t$-th step can be computed as:
\begin{equation}
\small
\label{eq:importance_metric}
  P^{l,(t)}_o = I^{l, (t)}_{o} + \gamma \frac{1}{|\mathcal{U}^{l,(t)}| - 1}\sum_{i \in \mathcal{U}^{l,(t)} \setminus o} I^{l, (t)}_{i|o},
\end{equation}
where $\mathcal{U}^{l,(t)}$ represents the remaining compression units (\emph{i.e.,} a set of not removed compression units) after $t-1$ steps and $\gamma$ is a hyper-parameter to trade-off the influence of removed units and remaining compression space.
Besides,
\begin{equation}
\footnotesize
  I^{l, (t)}_{o} = \mathrm{S}[(\mathcal{G}^l * (\overline{\mathcal{W}}^{l,(t)}_{o} - \mathcal{W}^l))^2]\ ,\  \overline{\mathcal{W}}^{l,(t)}_{o} = f(\overline{\mathcal{W}}^{l,(t-1)}, o),
\end{equation}
and
\begin{equation}
\small
  I^{l, (t)}_{i|o} = \mathrm{S}[(\mathcal{G}^l * (\overline{\mathcal{W}}^{l,(t)}_{i|o} - \mathcal{W}^l))^2], \  \overline{\mathcal{W}}^{l,(t)}_{i|o} = f(\overline{\mathcal{W}}^{l, (t)}_{o}, i).
\end{equation}
The first item in Eq.~\ref{eq:importance_metric} represents the importance of the $o$-th compression unit based on the compressed weight $\overline{\mathcal{W}}^{l, (t-1)}$ at the ($t-1$)-th step.
The second item represents the potential information loss for searching the remaining compression space, which is an average information loss of the weights generated by removing the remaining compression units of $\overline{\mathcal{W}}^{l, (t)}_{o}$.
Based on proposed important metric, we compute the importance of each compression unit and remove the least important one step-by-step.
The compression process stops when the compression rate of $\overline{\mathcal{W}}^{l, (t)}$ is larger than the target compression rate at the $l$-th layer.
After finishing the compression of all layers parallelly, the approximated weight $\overline{\mathcal{W}}^{l, (t)}$ will be transformed to $\widetilde{\mathcal{W}}^l$ via Eq.~\ref{eq:compressed_weight}.
Finally, fine-tuning is used to improve the accuracy of the compressed network.

However, the time complexity of the above computation process is $O((c^l+r^l)^3)$\footnote{We omit the time complexity of SVD and matrix multiplication to simplify the analysis.}, which significantly affects the efficiency of offline compression.
Instead, we propose to remove the $T$ compression units after measuring the importance metric each time to accelerate the compression process, where $T$ is a balance hyper-parameter.
As such, the computational complexity for the importance metric $P_o^{l,(t)}$ is $O(c^l+r^l)$, we re-formulate the second item in the importance metric as:
\begin{equation}
\label{eq:simple_1}
\footnotesize
\begin{aligned}
\gamma \frac{1}{|\mathcal{U}^{l,(t)}|-1} &\sum_{i \in \mathcal{U}^{l,(t)} \setminus o} I^{l, (t)}_{i|o} \\
  &= \gamma \frac{1}{|\mathcal{U}^{l,(t)}_{o}|}\sum_{i \in \mathcal{U}^{l,(t)}_{o}} \mathrm{S}[(\mathcal{G}^l * (\overline{\mathcal{W}}^{l,(t)}_{i|o} - \mathcal{W}^l))^2] \\
  &= \gamma \frac{1}{|\mathcal{U}^{l,(t)}_{o}|}\sum_{i \in \mathcal{U}^{l,(t)}_{o}} \mathrm{S}[(\mathcal{G}^l * (\theta^{l, (t)}_{i|o} + \theta^{l, (t)}_{o}))^2]
\end{aligned}
\end{equation}
where $\theta^{l, (t)}_{i|o} = \overline{\mathcal{W}}^{l,(t)}_{i|o} - \overline{\mathcal{W}}^{l, (t)}_{o}$ and $\theta^{l, (t)}_{o} = \overline{\mathcal{W}}^{l, (t)}_{o} - \mathcal{W}^l$, $*$ represents element-wise multiplication and $\mathcal{U}^{l,t}_{o} = \mathcal{U}^{l,t} \setminus o$.
The detailed computation process is presented in the supplementary materials.
Finally, we can compute the importance metric as:

\begin{equation}
\small
\begin{aligned}
  P^{l,(t)}_{o} 
  &= (1+\gamma)\mathrm{S}[(\mathcal{G}^l * (\overline{\mathcal{W}}^{l,(t)}_{o} - \mathcal{W}^l))^2] \\
  & - \gamma \frac{4}{|\mathcal{U}^{l,(t)}_{o}|} \mathrm{S}[(\mathcal{G}^l)^2 * \theta^{l, (t)}_{o} * \overline{\mathcal{W}}^{l, (t)}_{o}] \\
  &\ + \gamma \frac{1}{|\mathcal{U}^{l,(t)}_{o}|} \mathrm{S}[(\mathcal{G}^l*\overline{\mathcal{W}}^{l, (t)}_{o})^2] \\
  & + \gamma \frac{1}{|\mathcal{U}^{l,(t)}_{o}|} \mathrm{S}[(\mathcal{G}^l)^2 * \phi((U^{l,(t)}_{o})^2 (\Sigma^{l,(t)}_{o})^2 (V_{o}^{{l,(t)}^{\top}})^2)],
\end{aligned}
\end{equation}
where $U^{l,(t)}_{o}$, $\overline{\Sigma}^{l,(t)}_{o}$ and $V_{o}^{{l,(t)}^{\top}}$ is the SVD result of $\phi^{-1}(\overline{\mathcal{W}}^{l,(t)}_{o})$.
Thus, the computational complexity for the importance metric is reduced to $O(1)$, which significantly accelerates the calculation of the importance metric.
Our heuristic compression algorithm is summarized in Alg.~A in the supplementary materials.

\section{Experiments}

The proposed CC scheme is implemented using Pytorch \cite{paszke2019pytorch} and evaluated on two datasets, CIFAR-10 and ImageNet ILSVRC 2012.
The input images in CIFAR-10 are classified into 10 classes, whose sizes are all $32\times32$.
The training set contains 50,000 images and the test set contains 10,000 images.
ImageNet ILSVRC 2012 consists of 1.28 million training images and 50,000 validation images for testing over 1,000 classes.

\begin{table*}
  \small
\begin{center}
\begin{tabular}{|p{2cm}<{\centering}|p{3.2cm}<{\centering}|p{2cm}<{\centering}|p{2cm}<{\centering}|p{1.8cm}<{\centering}|p{1.8cm}<{\centering}|}
\hline
Model & Method & FLOPs (PR) & \#Param. (PR) & Top-1 Acc\% & Top-5 Acc\% \\
\hline
\multirow{5}*{VGG-16} & Baseline & 15.48B & 138M & 71.59 & 90.38\\

 & LRSD\cite{yu2017compressing} & - & 9.70M(93.0\%) & 68.75 & - \\
 
 & Coreset\cite{dubey2018coreset} & - & 9.81M(92.9\%) & 68.56 & - \\
 
 & Coreset\cite{dubey2018coreset} & - & 8.70M(93.7\%) & 68.16 & - \\
\cline{2-6}
& \textbf{CC-GAP ($C = 0.5$)} & \textbf{7.37B(52.4\%)} & \textbf{8.35M(93.9\%)} & \textbf{68.81} & \textbf{88.70}  \\

\hline

\multirow{6}*{ResNet-50} & Baseline & 4.10B & 25.56M & 76.15 & 92.87 \\

 & Hinge\cite{li2020group} & 1.91B(53.4\%) & - & 74.70 & - \\
 
 & Rethinking\cite{liu2018rethinking}+SVD\cite{zhang2015accelerating} & 1.93B(52.9\%) & - & 74.75 & 92.17 \\
 
 & FPGM \cite{he2019filter}+SVD\cite{zhang2015accelerating} & 1.93B(52.9\%) & - & 75.47 & 92.55 \\

\cline{2-6}
& \textbf{CC ($C = 0.5$)} & \textbf{1.93B(52.9\%)} & \textbf{13.2M(48.4\%)} & \textbf{75.59} & \textbf{92.64}  \\

& \textbf{CC ($C = 0.6$)} & \textbf{1.53B(62.7\%)} & \textbf{10.58M(58.6\%)} & \textbf{74.54} & \textbf{92.25} \\

\hline

\end{tabular}
\end{center}
\vspace{-2em}
\caption{Comparison with multiple compression operations-based methods on ImageNet2012.}
\vspace{-1em}
\label{tab:imagenet}
\end{table*}

\subsection{Experimental Settings}
All networks are trained via the stochastic gradient descent (SGD) with momentum 0.9.
On CIFAR-10, we train the networks for 300 epochs using the mini-batch size of 128.
The initial learning rate is set to 0.01 and is multiplied by 0.1 at 50\% and 75\% of the total epoch number.
On ImageNet, we train the networks for 90 and 120 epochs with the initial learning rate is set to 0.001 and 0.01 for VGG and ResNet, respectively.
Both mini-batch sizes on VGG and ResNets are set to 256.
The learning rate is divided by 10 at epoch 30, 60 and 90.
For fine-tuning the compressed networks, we follow the previous work\footnote{https://github.com/d-li14/mobilenetv2.pytorch} for training MobileNet-V2.
We do not compress the first and the last layers of the networks.
In our method, we use $T^l=\lfloor 0.01*(c^l+r^l) \rfloor$ to control the calculation interval of the importance metric and set $\gamma$ to $0.5$.

\subsection{Comparison with State-of-the-Art Methods}

\textbf{CIFAR-10.}
We compare our CC scheme with other methods with a single compression operation (either channel pruning or tensor decomposition) on ResNet-56, DenseNet-40, VGGNet, and GoogLeNet.
As shown in Table~\ref{tab:cifar10}, our method achieves the largest reductions of parameters and FLOPs, but with better performance, compared to other SOTA methods.
For instance, compared to HRank \cite{lin2020hrank} based on channel pruning, CC achieves higher reductions of FLOPs (59.9\% vs. 54.6\%) and parameters (63.3\% vs. 55.4\%) with higher accuracy (94.88\% vs. 94.53\%) on GoogLeNet.
Meanwhile, compared to ENC \cite{kim2019efficient} based on tensor decomposition, CC has better performance on ResNet-56 (52.0\% vs. 49.6\% in FLOPs reduction, and 93.64\% vs. 93.09\% in top-1 accuracy).

\textbf{ImageNet 2012.}
We further compare our CC scheme with other methods based on multiple compression operations on VGG-16 and ResNet-50.
``CC-GAP'' represents that the convolutional layers are compressed by CC and all the fully-connected (FC) layers in VGG-16 are replaced by a global average pooling (GAP) and one FC layer.
We also employ the same training parameters to fine-tune the compressed model by CC-GAP.
As shown in Table~\ref{tab:imagenet}, our method achieves the best performance with only a decrease of $0.56$\% in Top-1 accuracy by a factor of $1.94\times$ compression and $2.12\times$ theoretical speedup on ResNet-50.
The result of ``Rethinking+SVD'' is obtained by compressing the pruned network ``ThiNet ResNet50-70\% Scratch-B'' in \cite{liu2018rethinking} and then using SVD-decomposition \cite{zhang2015accelerating} to conduct further compression.
This method is regraded as the intuitive combination of two compression operations.
Meanwhile, the result of ``FPGM+SVD'' is also obtained by compressing the pruned network ``FPGM-only 30\%'' in \cite{he2019filter} and then performing SVD-decomposition \cite{zhang2015accelerating}.
More results are given in the supplementary materials.
In Table~\ref{tab:mobilenetv2}, we compare the proposed CC scheme with the state-of-the-art AutoML based pruning methods on MobileNet-V2.
We find that our method achieves the best performance with 70.91\% top-1 accuracy and only 215M FLOPs on ImageNet 2012.

\begin{table}[H]
\small
\begin{tabular}{|p{3.7cm}<{\centering}|p{1cm}<{\centering}|p{2.5cm}<{\centering}|}
\hline
 Method & FLOPs & Top-1 Acc\%  \\
\hline
 MobileNet-V2 & 300M & 71.88 \\

$0.75\times$ MobileNet-v2\cite{sandler2018mobilenetv2} & 220M & 69.80 \\

AMC\cite{he2018amc} & 220M & 70.80 \\

\hline
\textbf{CC($C=0.12$)} & \textbf{215M} & \textbf{70.91} \\

\hline

\end{tabular}
\vspace{-1em}
\caption{Results of MobileNet-V2 on ImageNet2012.}
\vspace{-1em}
\label{tab:mobilenetv2}
\end{table}
\begin{table}[H]
\small
\begin{tabular}{|p{1.35cm}<{\centering}|p{2cm}<{\centering}|p{1cm}<{\centering}|p{1cm}<{\centering}|p{1cm}<{\centering}|}
\hline
 \multirow{2}*{Model} & \multirow{2}*{Method} & \multicolumn{3}{c|}{Batch Size}  \\
 \cline{3-5}
 ~& ~ & 1 & 8 & 32 \\
 \hline
 \multirow{2}*{VGG-16} & Baseline & 225ms & 1224ms & 4603ms \\
 \cline{2-5}
 ~ & \textbf{CC($C=0.5$)} & \textbf{173ms} & \textbf{784ms} & \textbf{2983ms} \\
 \hline
 \multirow{2}*{ResNet-50} & Baseline & 96ms & 584ms & 2420ms \\
 \cline{2-5}
 ~ & \textbf{CC($C=0.5$)} & \textbf{88ms} & \textbf{420ms} & \textbf{1776ms}  \\
 \hline
\end{tabular}
\vspace{-1em}
\caption{Results of latency on CPU.}
\vspace{-2em}
\label{tab:speed}
\end{table}
Moreover, we demonstrate the effectiveness of our method in wall-clock time speedup using VGG-16 and ResNet-50 on PyTorch using the CPU AMD Ryzen Threadripper 1900X.
As shown in Table~\ref{tab:speed}, our method achieves $1.54\times$ and $1.36\times$ acceleration rates with batch size 32 on VGG-16 and ResNet-50, respectively.

\begin{figure}[H]
\centering
  \includegraphics[width=0.8\columnwidth]{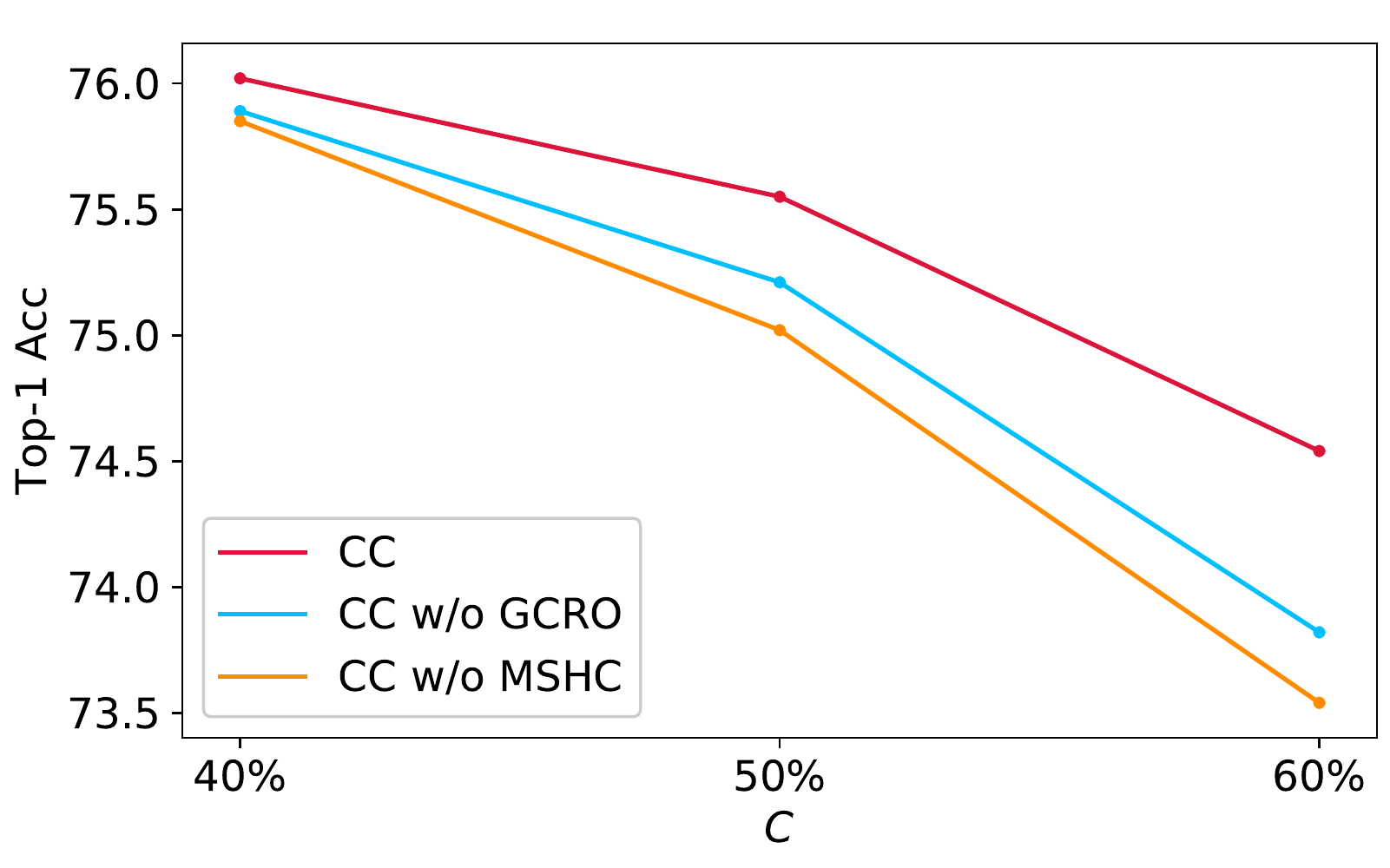}
\vspace{-1em}
\caption{Ablation study of CC with ResNet-50 on ImageNet 2012.}
\vspace{-1em}
      \label{fig:method_analyse}
\end{figure}

\subsection{Ablation Study}
 \vspace{-0.5em}
\subsubsection{Effect of the Two Components in CC}
\vspace{-0.7em}
In this section, we evaluate the effectiveness of CC's Global Compression Rate Optimization (GCRO) and Multi-Step Heuristic Compression (MSHC).
We use the same compression rate for each layer to replace GCRO (\emph{i.e.,} CC w/o GCRO), and only compute the importance metric once by setting $T = r^l+c^l$ to replace MSHC (\emph{i.e.,} CC w/o MSHC), separately.
As shown in Fig.~\ref{fig:method_analyse}, both GCRO and MSHC indeed increase the performance of the compressed network.
It demonstrates the effectiveness of GCRO and MSHC.
We also observe that MSHC is more important in our CC scheme, compared to GCRO.
This reveals the compression strategy is more important than the compression rate decision in the union compression.

 \vspace{-1em}
\subsubsection{Effect of the Compression Order}
\begin{figure}[H]
\hspace{1.5em}
\vspace{-1em}
   \includegraphics[width=0.8\columnwidth]{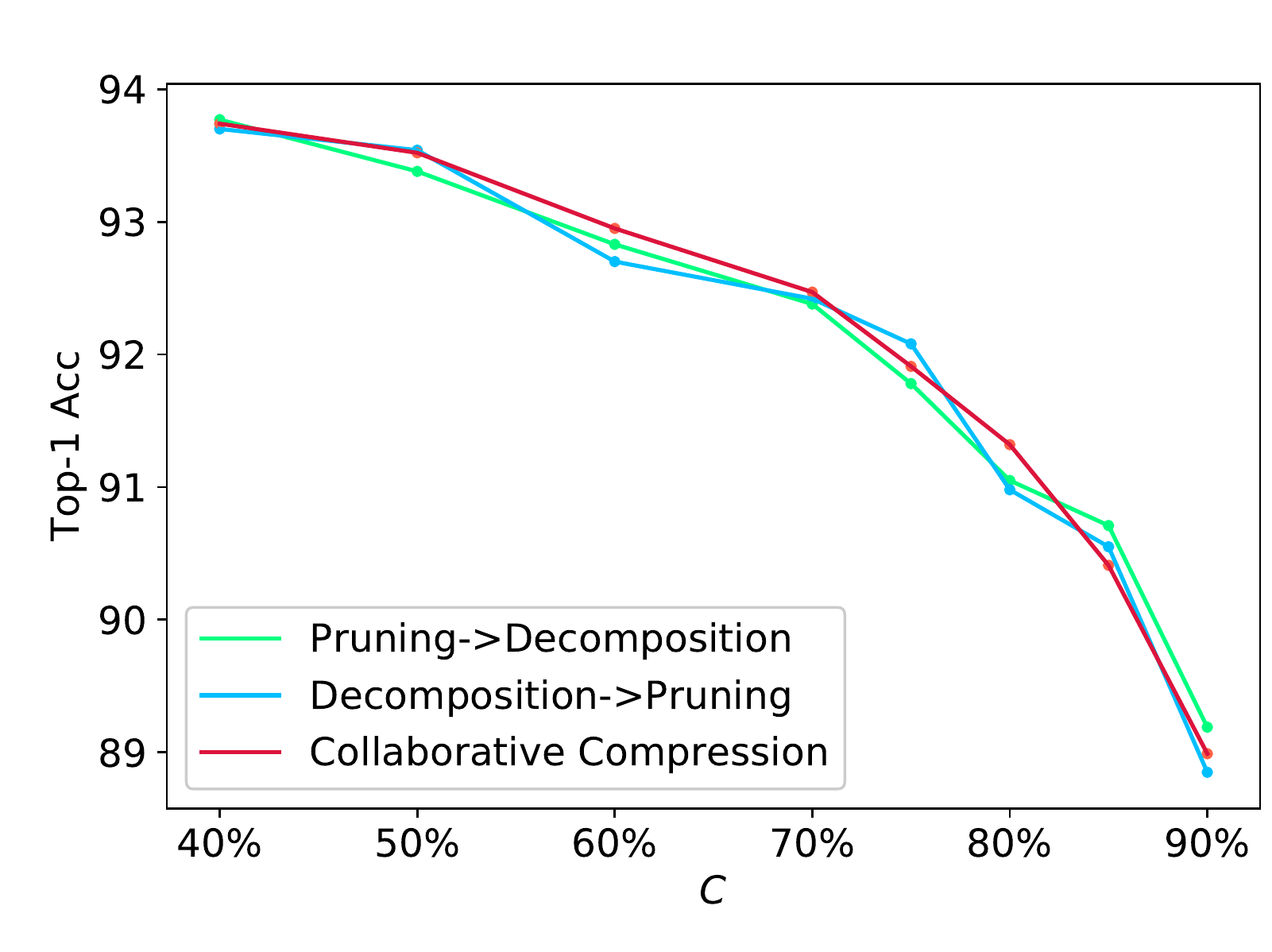}
\vspace{-1em}
\caption{Comparison of different variants of CC with ResNet-56 on CIFAR-10.}
\vspace{-1em}
      \label{fig:compress_process}
\end{figure}

\vspace{-0.7em}
We propose two different variants of our method to investigate the effect of the compression order for the performance of the compressed network under a fixed structure.
Our CC method simultaneously prunes and decomposes the weights during compression.
Therefore, we can obtain the architecture of the compressed network, including the pruned $t_1$ input channels and $t_2$ singular values. 
In Fig.~\ref{fig:compress_process}, ``Pruning$\to$Decomposition'' represents that we first prune the weights and then fine-tune the pruned network, and continue to decompose the pruned network with the following fine-tuning.
``Decomposition$\to$Pruning'' is in reverse order to ``Pruning$\to$Decomposition’’.
Note that these two methods can obtain the same compressed network structure (i.e. the same $t_1$ and $t_2$) as that by our CC scheme, but only with different compression orders.
The result demonstrates that the compression order (\emph{i.e.,} ``Pruning$\to$Decomposition'', ``Decomposition$\to$Pruning'', and Collaborative Compression) has less effect on the final performance.
In other words, the different compression strategies influence the performance of the compressed network by producing different network's structures rather than the network's weights.
The result demonstrates that the network's structure is more important than network's weights in the compression algorithm, which is also shown in \cite{liu2018rethinking}.

\begin{figure}[H]
\centering
\includegraphics[width=0.8\columnwidth]{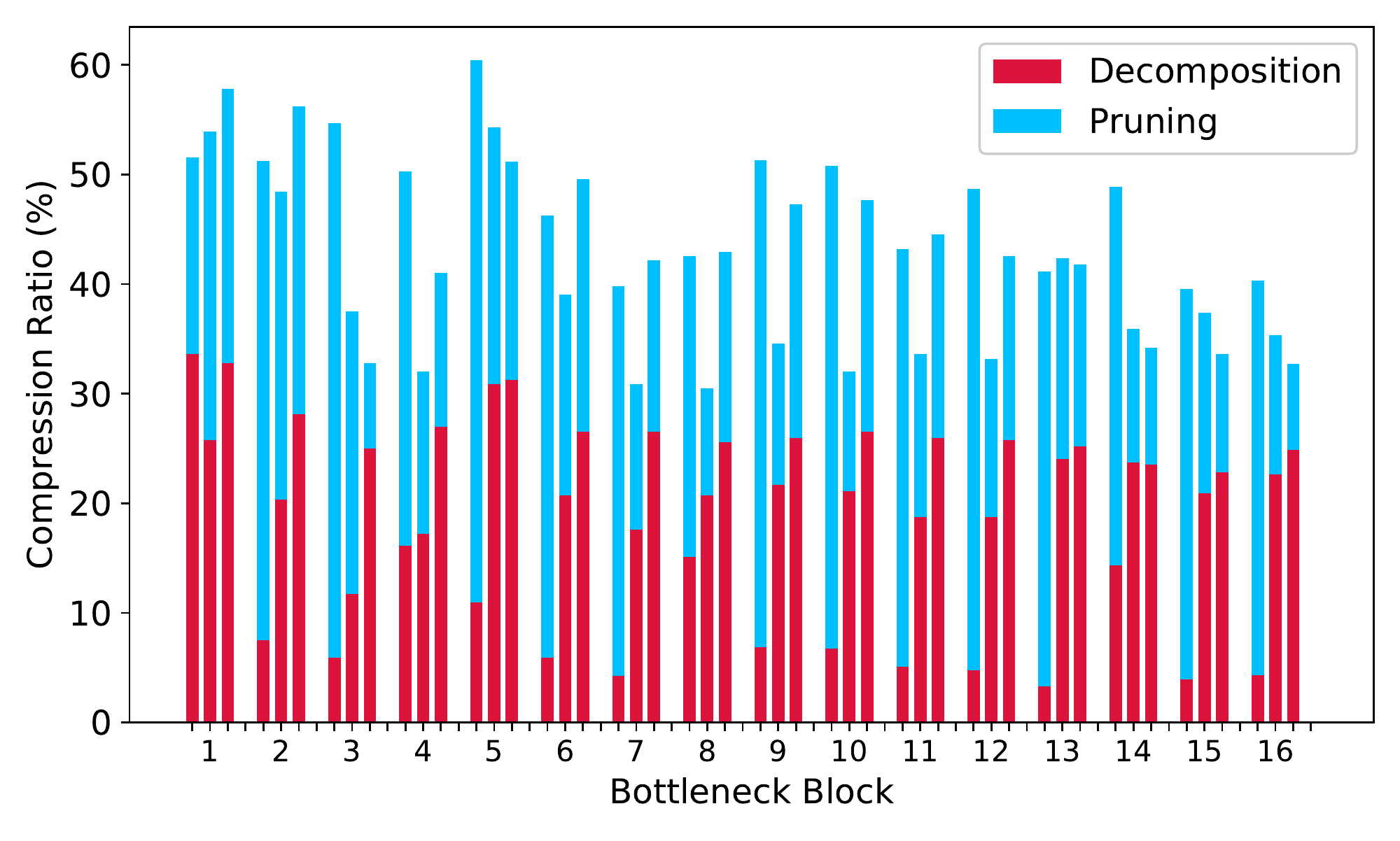}
\vspace{-1em}
\caption{Compression rates of different compression operations in different layers of ResNet-50.}
\vspace{-1em}
      \label{fig:co_analysis}
\end{figure}

\vspace{-1em}
\subsubsection{Compression Percentage Analysis.}
\vspace{-0.7em}
Fig.~\ref{fig:co_analysis} shows the proportion of removed compression units in channel pruning and tensor-decomposition from different layers in ResNet-50.
Each bottleneck block has three convolutional layers.
In our CC, the first convolution layer in each bottleneck block tends to generate sparse weights rather than low-rank weights.
It reveals that the diversity of input features is more important for each bottleneck block in ResNet-50.
In contrast, the final layer of the block pays more attention to produce compact features.

\section{Conclusion}

In this paper, we first investigate the problem of post-training union compression and propose a novel unified compression framework \textit{Collaborative Compression} (CC) for CNN.
Our method is divided into two stages: Global Compression Rate Optimization and Multi-Step Heuristic Compression.
The first stage decides the compression rates of the layers by solving an optimization problem based on their compression sensitivity.
After that, we compress each layer separately by removing the less important compression units step-by-step.
Therefore, our compression process is more reliable, which fully takes the effect of the removed compression units and remaining compression space into account.
Extensive experiments on various modern CNNs demonstrate the effectiveness of CC for reducing the computational complexity and model size.

\vspace{-0.5em}
\section*{Acknowledgments}
\vspace{-0.5em}
This work is supported by the National Science Fund for Distinguished Young Scholars (No.62025603), the National Natural Science Foundation of China (No.U1705262, No.62072386, No.62072387, No.62072389, No.62002305, No.61772443, No.61802324 and No.61702136) and Guangdong Basic and Applied Basic Research Foundation(No.2019B1515120049).

{\small
\bibliographystyle{ieee_fullname}
\bibliography{egbib}

\begin{thebibliography}{10}\itemsep=-1pt

\bibitem{chen2018encoder}
Liang-Chieh Chen, Yukun Zhu, George Papandreou, Florian Schroff, and Hartwig
  Adam.
\newblock Encoder-decoder with atrous separable convolution for semantic image
  segmentation.
\newblock {\em European Conference on Computer Vision (ECCV)}, 2018.

\bibitem{ding2019centripetal}
Xiaohan Ding, Guiguang Ding, Yuchen Guo, and Jungong Han.
\newblock Centripetal sgd for pruning very deep convolutional networks with
  complicated structure.
\newblock {\em IEEE Conference on Computer Vision and Pattern Recognition
  (CVPR)}, 2019.

\bibitem{ding2019approximated}
Xiaohan Ding, Guiguang Ding, Yuchen Guo, Jungong Han, and Chenggang Yan.
\newblock Approximated oracle filter pruning for destructive cnn width
  optimization.
\newblock {\em International Conference on Machine Learning}, 2019.

\bibitem{dubey2018coreset}
Abhimanyu Dubey, Moitreya Chatterjee, and Narendra Ahuja.
\newblock Coreset-based neural network compression.
\newblock {\em European Conference on Computer Vision (ECCV)}, 2018.

\bibitem{han2015deep}
Song Han, Huizi Mao, and William~J Dally.
\newblock Deep compression: Compressing deep neural networks with pruning,
  trained quantization and huffman coding.
\newblock {\em International Conference on Learning Representations (ICLR)},
  2016.

\bibitem{han2015learning}
Song Han, Jeff Pool, John Tran, and William Dally.
\newblock Learning both weights and connections for efficient neural network.
\newblock {\em Advances in Neural Information Processing Systems (NeurIPS)},
  2015.

\bibitem{he2016deep}
Kaiming He, Xiangyu Zhang, Shaoqing Ren, and Jian Sun.
\newblock Deep residual learning for image recognition.
\newblock {\em IEEE Conference on Computer Vision and Pattern Recognition
  (CVPR)}, 2016.

\bibitem{he2018amc}
Yihui He, Ji Lin, Zhijian Liu, Hanrui Wang, Li-Jia Li, and Song Han.
\newblock Amc: Automl for model compression and acceleration on mobile devices.
\newblock {\em European Conference on Computer Vision (ECCV)}, 2018.

\bibitem{he2019filter}
Yang He, Ping Liu, Ziwei Wang, Zhilan Hu, and Yi Yang.
\newblock Filter pruning via geometric median for deep convolutional neural
  networks acceleration.
\newblock {\em IEEE Conference on Computer Vision and Pattern Recognition
  (CVPR)}, 2019.

\bibitem{he2017channel}
Yihui He, Xiangyu Zhang, and Jian Sun.
\newblock Channel pruning for accelerating very deep neural networks.
\newblock {\em International Conference on Computer Vision (ICCV)}, 2017.

\bibitem{hu2016network}
Hengyuan Hu, Rui Peng, Yu-Wing Tai, and Chi-Keung Tang.
\newblock Network trimming: A data-driven neuron pruning approach towards
  efficient deep architectures.
\newblock {\em arXiv preprint arXiv:1607.03250}, 2016.

\bibitem{huang2018condensenet}
Gao Huang, Shichen Liu, Laurens Van~der Maaten, and Kilian~Q Weinberger.
\newblock Condensenet: An efficient densenet using learned group convolutions.
\newblock {\em IEEE Conference on Computer Vision and Pattern Recognition
  (CVPR)}, 2018.

\bibitem{jacob2017quantization}
Benoit Jacob, Skirmantas Kligys, Bo Chen, Menglong Zhu, Matthew Tang, Andrew
  Howard, Hartwig Adam, and Dmitry Kalenichenko.
\newblock Quantization and training of neural networks for efficient
  integer-arithmetic-only inference.
\newblock {\em IEEE Conference on Computer Vision and Pattern Recognition
  (CVPR)}, 2018.

\bibitem{kim2019efficient}
Hyeji Kim, Muhammad Umar~Karim Khan, and Chong-Min Kyung.
\newblock Efficient neural network compression.
\newblock {\em IEEE Conference on Computer Vision and Pattern Recognition
  (CVPR)}, 2019.

\bibitem{kim2015compression}
Yong-Deok Kim, Eunhyeok Park, Sungjoo Yoo, Taelim Choi, Lu Yang, and Dongjun
  Shin.
\newblock Compression of deep convolutional neural networks for fast and low
  power mobile applications.
\newblock {\em International Conference on Learning Representations (ICLR)},
  2015.

\bibitem{krizhevsky2012imagenet}
Alex Krizhevsky, Ilya Sutskever, and Geoffrey~E Hinton.
\newblock Imagenet classification with deep convolutional neural networks.
\newblock {\em Advances in Neural Information Processing Systems (NeurIPS)},
  2012.

\bibitem{lebedev2014speeding}
Vadim Lebedev, Yaroslav Ganin, Maksim Rakhuba, Ivan Oseledets, and Victor
  Lempitsky.
\newblock Speeding-up convolutional neural networks using fine-tuned
  cp-decomposition.
\newblock {\em International Conference on Learning Representations (ICLR)},
  2014.

\bibitem{li2018constrained}
Chong Li and CJ Richard~Shi.
\newblock Constrained optimization based low-rank approximation of deep neural
  networks.
\newblock {\em European Conference on Computer Vision (ECCV)}, 2018.

\bibitem{li2016pruning}
Hao Li, Asim Kadav, Igor Durdanovic, Hanan Samet, and Hans~Peter Graf.
\newblock Pruning filters for efficient convnets.
\newblock {\em International Conference on Learning Representations (ICLR)},
  2016.

\bibitem{li2020group}
Yawei Li, Shuhang Gu, Christoph Mayer, Luc Van~Gool, and Radu Timofte.
\newblock Group sparsity: The hinge between filter pruning and decomposition
  for network compression.
\newblock {\em IEEE Conference on Computer Vision and Pattern Recognition
  (CVPR)}, 2020.

\bibitem{lin2020hrank}
Mingbao Lin, Rongrong Ji, Yan Wang, Yichen Zhang, Baochang Zhang, Yonghong
  Tian, and Ling Shao.
\newblock Hrank: Filter pruning using high-rank feature map.
\newblock {\em IEEE Conference on Computer Vision and Pattern Recognition
  (CVPR)}, 2020.

\bibitem{lin2018holistic}
Shaohui Lin, Rongrong Ji, Chao Chen, Dacheng Tao, and Jiebo Luo.
\newblock Holistic cnn compression via low-rank decomposition with knowledge
  transfer.
\newblock {\em IEEE transactions on pattern analysis and machine intelligence
  (TPAMI)}, 2018.

\bibitem{lin2019toward}
Shaohui Lin, Rongrong Ji, Yuchao Li, Cheng Deng, and Xuelong Li.
\newblock Toward compact convnets via structure-sparsity regularized filter
  pruning.
\newblock {\em IEEE transactions on neural networks and learning systems
  (TNNLS)}, 2019.

\bibitem{lin2018accelerating}
Shaohui Lin, Rongrong Ji, Yuchao Li, Yongjian Wu, Feiyue Huang, and Baochang
  Zhang.
\newblock Accelerating convolutional networks via global \& dynamic filter
  pruning.
\newblock {\em IJCAI}, pages 2425--2432, 2018.

\bibitem{lin2019towards}
Shaohui Lin, Rongrong Ji, Chenqian Yan, Baochang Zhang, Liujuan Cao, Qixiang
  Ye, Feiyue Huang, and David Doermann.
\newblock Towards optimal structured cnn pruning via generative adversarial
  learning.
\newblock {\em IEEE Conference on Computer Vision and Pattern Recognition
  (CVPR)}, 2019.

\bibitem{liu2017learning}
Zhuang Liu, Jianguo Li, Zhiqiang Shen, Gao Huang, Shoumeng Yan, and Changshui
  Zhang.
\newblock Learning efficient convolutional networks through network slimming.
\newblock {\em International Conference on Computer Vision (ICCV)}, 2017.

\bibitem{liu2018rethinking}
Zhuang Liu, Mingjie Sun, Tinghui Zhou, Gao Huang, and Trevor Darrell.
\newblock Rethinking the value of network pruning.
\newblock {\em International Conference on Learning Representations (ICLR)},
  2018.

\bibitem{luo2017thinet}
Jian-Hao Luo, Jianxin Wu, and Weiyao Lin.
\newblock Thinet: A filter level pruning method for deep neural network
  compression.
\newblock {\em International Conference on Computer Vision (ICCV)}, 2017.

\bibitem{ma2019unified}
Yuzhe Ma, Ran Chen, Wei Li, Fanhua Shang, Wenjian Yu, Minsik Cho, and Bei Yu.
\newblock A unified approximation framework for compressing and accelerating
  deep neural networks.
\newblock {\em 2019 IEEE 31st International Conference on Tools with Artificial
  Intelligence (ICTAI)}, pages 376--383, 2019.

\bibitem{molchanov2019importance}
Pavlo Molchanov, Arun Mallya, Stephen Tyree, Iuri Frosio, and Jan Kautz.
\newblock Importance estimation for neural network pruning.
\newblock {\em IEEE Conference on Computer Vision and Pattern Recognition
  (CVPR)}, 2019.

\bibitem{molchanov2016pruning}
Pavlo Molchanov, Stephen Tyree, Tero Karras, Timo Aila, and Jan Kautz.
\newblock Pruning convolutional neural networks for resource efficient transfer
  learning.
\newblock {\em arXiv preprint arXiv:1611.06440}, 3, 2016.

\bibitem{paszke2019pytorch}
Adam Paszke, Sam Gross, Francisco Massa, Adam Lerer, James Bradbury, Gregory
  Chanan, Trevor Killeen, Zeming Lin, Natalia Gimelshein, Luca Antiga, et~al.
\newblock Pytorch: An imperative style, high-performance deep learning library.
\newblock {\em Advances in Neural Information Processing Systems (NeurIPS)},
  2019.

\bibitem{Redmon2018YOLOv3}
Joseph Redmon and Ali Farhadi.
\newblock Yolov3: An incremental improvement.
\newblock 2018.

\bibitem{Ren2015Faster}
Shaoqing Ren, Kaiming He, Ross Girshick, and Jian Sun.
\newblock Faster r-cnn: towards real-time object detection with region proposal
  networks.
\newblock {\em Advances in Neural Information Processing Systems (NeurIPS)},
  2015.

\bibitem{sandler2018mobilenetv2}
Mark Sandler, Andrew Howard, Menglong Zhu, Andrey Zhmoginov, and Liang-Chieh
  Chen.
\newblock Mobilenetv2: Inverted residuals and linear bottlenecks.
\newblock {\em IEEE Conference on Computer Vision and Pattern Recognition
  (CVPR)}, 2018.

\bibitem{simonyan2014very}
Karen Simonyan and Andrew Zisserman.
\newblock Very deep convolutional networks for large-scale image recognition.
\newblock {\em arXiv preprint arXiv:1409.1556}, 2014.

\bibitem{tung2018clip}
Frederick Tung and Greg Mori.
\newblock Clip-q: Deep network compression learning by in-parallel
  pruning-quantization.
\newblock {\em IEEE Conference on Computer Vision and Pattern Recognition
  (CVPR)}, 2018.

\bibitem{veit2018convolutional}
Andreas Veit and Serge Belongie.
\newblock Convolutional networks with adaptive inference graphs.
\newblock {\em European Conference on Computer Vision (ECCV)}, 2018.

\bibitem{veit2016residual}
Andreas Veit, Michael~J Wilber, and Serge Belongie.
\newblock Residual networks behave like ensembles of relatively shallow
  networks.
\newblock {\em Advances in Neural Information Processing Systems (NeurIPS)},
  2016.

\bibitem{wen2016learning}
Wei Wen, Chunpeng Wu, Yandan Wang, Yiran Chen, and Hai Li.
\newblock Learning structured sparsity in deep neural networks.
\newblock {\em Advances in Neural Information Processing Systems (NeurIPS)},
  2016.

\bibitem{wen2017coordinating}
Wei Wen, Cong Xu, Chunpeng Wu, Yandan Wang, Yiran Chen, and Hai Li.
\newblock Coordinating filters for faster deep neural networks.
\newblock {\em International Conference on Computer Vision (ICCV)}, 2017.

\bibitem{yu2018nisp}
Ruichi Yu, Ang Li, Chun-Fu Chen, Jui-Hsin Lai, Vlad~I Morariu, Xintong Han,
  Mingfei Gao, Ching-Yung Lin, and Larry~S Davis.
\newblock Nisp: Pruning networks using neuron importance score propagation.
\newblock {\em IEEE Conference on Computer Vision and Pattern Recognition
  (CVPR)}, 2018.

\bibitem{yu2017compressing}
Xiyu Yu, Tongliang Liu, Xinchao Wang, and Dacheng Tao.
\newblock On compressing deep models by low rank and sparse decomposition.
\newblock {\em IEEE Conference on Computer Vision and Pattern Recognition
  (CVPR)}, 2017.

\bibitem{zhang2015accelerating}
Xiangyu Zhang, Jianhua Zou, Kaiming He, and Jian Sun.
\newblock Accelerating very deep convolutional networks for classification and
  detection.
\newblock {\em IEEE transactions on pattern analysis and machine intelligence
  (TPAMI)}, 38(10):1943--1955, 2015.

\bibitem{zhou2016dorefa}
Shuchang Zhou, Yuxin Wu, Zekun Ni, Xinyu Zhou, He Wen, and Yuheng Zou.
\newblock Dorefa-net: Training low bitwidth convolutional neural networks with
  low bitwidth gradients.
\newblock {\em IEEE Conference on Computer Vision and Pattern Recognition
  (CVPR)}, 2016.

\end{thebibliography}
}

\end{multicols}

\newpage
\appendix


\section{Derivation of Equation (14) in the Paper}

The important metric in our method is:
\begin{equation}
\small
\label{eq:importance_metric_new}
  P^{l,(t)}_o = I^{l, (t)}_{o} + \gamma \frac{1}{|\mathcal{U}^{l,(t)}| - 1}\sum_{i \in \mathcal{U}^{l,(t)} \setminus o} I^{l, (t)}_{i|o}.
\end{equation}
The second item in it can be re-formulated as:
\begin{equation}
\label{eq:simple_1_new}
\small
\begin{aligned}
\gamma \frac{1}{|\mathcal{U}^{l,(t)}|-1}\sum_{i \in \mathcal{U}^{l,(t)} \setminus o} I^{l,(t)}_{i|o} 
  &= \gamma \frac{1}{|\mathcal{U}^{l,(t)}_{o}|}\sum_{i \in \mathcal{U}^{l,(t)}_{o}} \mathrm{S}[(\mathcal{G}^l * (\overline{\mathcal{W}}^{l,(t)}_{i|o} - \mathcal{W}^l)^2] \\
  &= \gamma \frac{1}{|\mathcal{U}^{l,(t)}_{o}|}\sum_{i \in \mathcal{U}^{l,(t)}_{o}} \mathrm{S}[(\mathcal{G}^l * (\overline{\mathcal{W}}^{l,(t)}_{i|o} - \overline{\mathcal{W}}^{l,(t)}_{o} + \overline{\mathcal{W}}^{l,(t)}_{o} - \mathcal{W}^l))^2]. \\
  &= \gamma \frac{1}{|\mathcal{U}^{l,(t)}_{o}|}\sum_{i \in \mathcal{U}^{l,(t)}_{o}}\mathrm{S}[(\mathcal{G}^l * (\theta^{l,(t)}_{i|o} + \theta^{l,(t)}_{o}))^2] \\
  &= \gamma \frac{1}{|\mathcal{U}^{l,(t)}_{o}|}\sum_{i \in \mathcal{U}^{l,(t)}_{o}} \mathrm{S} [(\mathcal{G}^l * \theta^{l,(t)}_{i|o})^2 + 2\mathcal{G}^l*\theta^{l,(t)}_{i|o} *\mathcal{G}^l * \theta^{l,(t)}_{o} + (\mathcal{G}^l * \theta^{l,(t)}_{o})^2] \\
  &=\gamma \frac{1}{|\mathcal{U}^{l,(t)}_{o}|} \mathrm{S}[(\mathcal{G}^l)^2 * \sum_{i \in \mathcal{U}^{l,(t)}_{o}} (\theta^{l,(t)}_{i|o})^2] \\
  & \ \ \ +\gamma \frac{2}{|\mathcal{U}^{l,(t)}|} \mathrm{S}[(\mathcal{G}^l)^2 * \theta^{l,(t)}_{o} * \sum_{i \in \mathcal{U}^{l,(t)} \setminus o} \theta^{l,(t)}_{i|o}] + \gamma \mathrm{S}[(\mathcal{G}^l * \theta^{l,(t)}_{o})^2],
\end{aligned}
\end{equation}
where $\theta^{l,(t)}_{i|o} = \overline{\mathcal{W}}^{l,(t)}_{i|o} - \overline{\mathcal{W}}^{l,(t)}_{o}$, $\theta^{l,(t)}_{o} = \overline{\mathcal{W}}^{l,(t)}_{o} - \mathcal{W}^l$.
$*$ represents element-wise multiplication, and $\mathcal{U}^{l,(t)}_{o} = \mathcal{U}^{l,(t)} \setminus o$.
Then, we consider the compression units in channel pruning $\mathcal{U}^{l,(t)}_{cp|o}$ and tensor decomposition $\mathcal{U}^{l,(t)}_{td|o}$ separately, where $\mathcal{U}^{l,(t)}_{cp|o} \cup \mathcal{U}^{l,(t)}_{td|o} = \mathcal{U}^{l,(t)}_{o}$.
For the remaining compression units of channel pruning (\emph{i.e.,} input channels), the first item in Eq.~\ref{eq:simple_1_new} can be rewritten as:
\begin{equation}
\small
  \gamma \frac{1}{|\mathcal{U}^{l,(t)}_{o}|} \mathrm{S}[(\mathcal{G}^l)^2 * \sum_{i \in \mathcal{U}^{l,(t)}_{cp|o}} (\theta^{l,(t)}_{i|o})^2] = \gamma \frac{1}{|\mathcal{U}^{l,(t)}_{o}|} \mathrm{S}[(\mathcal{G}^l)^2 * (-\overline{\mathcal{W}}^{l,(t)}_{o})^2] =\gamma \frac{1}{|\mathcal{U}^{l,(t)}_{o}|} \mathrm{S}[(\mathcal{G}^l*\overline{\mathcal{W}}^{l,(t)}_{o})^2],
\end{equation}
and the second item can also be rewritten as:
\begin{equation}
\small
  \gamma \frac{2}{|\mathcal{U}^{l,(t)}_{o}|} \mathrm{S}[(\mathcal{G}^l)^2 * \theta^{l,(t)}_{o} * \sum_{i \in \mathcal{U}^{l,(t)}_{cp|o}} \theta^{l,(t)}_{i|o}] = -\gamma \frac{2}{|\mathcal{U}^{l,(t)}_{o}|} \mathrm{S}[(\mathcal{G}^l)^2 * \theta^{l,(t)}_{o} * \overline{\mathcal{W}}^{l,(t)}_{o}].
\end{equation}
Therefore, the second item in Eq.~\ref{eq:importance_metric_new} for channel pruning becomes:
\begin{equation}
\small
  \gamma \frac{1}{|\mathcal{U}^{l,(t)}_{o}|}\sum_{i \in \mathcal{U}^{l,(t)}_{cp|o}} I^{l,(t)}_{i|o} = \gamma \frac{1}{|\mathcal{U}^{l,(t)}_{o}|}  \mathrm{S}[(\mathcal{G}^l*\overline{\mathcal{W}}^{l,(t)}_{o})^2] - \gamma \frac{2}{|\mathcal{U}^{l,(t)}_{o}|}  \mathrm{S}[(\mathcal{G}^l)^2 * \theta^{l,(t)}_{o} * \overline{\mathcal{W}}^{l,(t)}_{o}] + \gamma \frac{|\mathcal{U}^{l,(t)}_{cp|o}|}{|\mathcal{U}^{l,(t)}_{o}|} \mathrm{S}[(\mathcal{G}^l * \theta^{l,(t)}_{o})^2].
\end{equation}
For the remaining compression units of tensor decomposition (\emph{i.e.,} singular values), the first item in Eq.~\ref{eq:simple_1} can be rewritten as:
\begin{equation}
\begin{aligned}
\small
\gamma \frac{1}{|\mathcal{U}^{l,(t)}_{o}|} \mathrm{S}[(\mathcal{G}^l)^2 * \sum_{i \in \mathcal{U}^{l,(t)}_{td|o}} (\theta^{l,(t)}_{i|o})^2]
&= \gamma \frac{1}{|\mathcal{U}^{l,(t)}_{o}|} \mathrm{S}[(\mathcal{G}^l)^2 * \sum_{i \in \mathcal{U}^{l,(t)}_{td|o}} \phi(-U^{l,(t)}_{:,i|o} \Sigma^{l,(t)}_{i,i|o} V^{{l,(t)}^{\top}}_{i,:|o})^2] \\
&= \gamma \frac{1}{|\mathcal{U}^{l,(t)}_{o}|} \mathrm{S}[(\mathcal{G}^l)^2 * \sum_{i \in \mathcal{U}^{l,(t)}_{td|o}} \phi((U^{l,(t)}_{:,i|o})^2 (\Sigma^{l,(t)}_{i,i|o})^2 (V^{{l,(t)}^{\top}}_{i,:|o})^2)] \\
&= \gamma \frac{1}{|\mathcal{U}^{l,(t)}_{o}|} \mathrm{S}[(\mathcal{G}^l)^2 * \phi((U^{l,(t)}_{o})^2 (\Sigma^{l,(t)}_{o})^2 (V_{o}^{{l,(t)}^{\top}})^2)], \\
\end{aligned}
\end{equation}
where $\overline{\mathcal{W}}^{l,(t)}_{o} = \phi(U^{l,(t)}_{o} \overline{\Sigma}^{l,(t)}_{o} V_{o}^{{l,(t)}^{\top}})$ is the SVD of $\phi^{-1}(\overline{\mathcal{W}}^{l,(t)}_{o})$.
The second item of Eq.2 can also be rewritten as:
\begin{equation}
\begin{aligned}
\small
  \gamma \frac{2}{|\mathcal{U}^{l,(t)}_{o}|} \mathrm{S}[(\mathcal{G}^l)^2 * \theta^{l,(t)}_{o} * \sum_{i \in \mathcal{U}^{l,(t)}_{td|o}} \theta^{l,(t)}_{i|o}] 
& = \gamma \frac{2}{|\mathcal{U}^{l,(t)}_{o}|} \mathrm{S}[(\mathcal{G}^l)^2 * \theta^{l,(t)}_{o} * \sum_{i \in \mathcal{U}^{l,(t)}_{td|o}} \phi(-U^{l,(t)}_{:,i|o} \Sigma^{l,(t)}_{i,i|o} V^{{l,(t)}^{\top}}_{i,:|o})] \\
& = -\gamma \frac{2}{|\mathcal{U}^{l,(t)}_{o}|} \mathrm{S}[(\mathcal{G}^l)^2 * \theta^{l,(t)}_{o} * \phi(U^{l,(t)}_{o} \Sigma^{l,(t)}_{o} V_{o}^{{l,(t)}^{\top}})] \\
& = -\gamma \frac{2}{|\mathcal{U}^{l,(t)}_{o}|} \mathrm{S}[(\mathcal{G}^l)^2 * \theta^{l,(t)}_{o} * \overline{\mathcal{W}}^{l,(t)}_{o}].
\end{aligned}
\end{equation}
Therefore, for the tensor decomposition, the second item in the importance metric Eq.~\ref{eq:importance_metric_new} is:
\begin{equation}
\begin{aligned}
\small
  \gamma \frac{1}{|\mathcal{U}^{l,(t)}_{o}|}\sum_{i \in \mathcal{U}^{l,(t)}_{td|o}} I^{l,(t)}_{i|o} = 
  &\gamma \frac{1}{|\mathcal{U}^{l,(t)}_{o}|} \mathrm{S}[(\mathcal{G}^l)^2 * \phi((U^{l,(t)}_{o})^2 (\Sigma^{l,(t)}_{o})^2 (V_{o}^{{l,(t)}^{\top}})^2)]  \\
  &- \gamma \frac{2}{|\mathcal{U}^{l,(t)}_{o}|} \mathrm{S}[(\mathcal{G}^l)^2 * \theta^{l,(t)}_{o} * \overline{\mathcal{W}}^{l,(t)}_{o}] + \gamma\frac{|\mathcal{U}^{l,(t)}_{td|o}|}{|\mathcal{U}^{l,(t)}_{o}|}\mathrm{S}[(\mathcal{G}^l * \theta^{l,(t)}_{o})^2].
\end{aligned}
\end{equation}
Finally, the importance metric can be formulated as follows:
\begin{equation}
\small
\label{eq:new_importance_metric}
\begin{aligned}
  P^{l,(t)}_{o} 
  & = I^{l,(t)}_{o} + \gamma \frac{1}{|\mathcal{U}^{l,(t)}_{o}|}\sum_{i \in \mathcal{U}^{l,(t)}_{cp|o}} I^{l,(t)}_{i|o} + \gamma \frac{1}{|\mathcal{U}^{l,(t)}_{o}|}\sum_{i \in \mathcal{U}^{l,(t)}_{td|o}} I^{l,(t)}_{i|o} \\ 
  &= (1+\gamma)\mathrm{S}[(\mathcal{G}^l * (\overline{\mathcal{W}}^{l,(t)}_{o} - \mathcal{W}^l))^2] - \gamma \frac{4}{|\mathcal{U}^{l,(t)}_{o}|} \mathrm{S}[(\mathcal{G}^l)^2 * \theta^{l, (t)}_{o} * \overline{\mathcal{W}}^{l, (t)}_{o}] \\
  &\ + \gamma \frac{1}{|\mathcal{U}^{l,(t)}_{o}|} \mathrm{S}[(\mathcal{G}^l*\overline{\mathcal{W}}^{l, (t)}_{o})^2] + \gamma \frac{1}{|\mathcal{U}^{l,(t)}_{o}|} \mathrm{S}[(\mathcal{G}^l)^2 * \phi((U^{l,(t)}_{o})^2 (\Sigma^{l,(t)}_{o})^2 (V_{o}^{{l,(t)}^{\top}})^2)].
\end{aligned}
\end{equation}

   

\section{Visualization of the Compression Process}

During the compression process, the approximated weight $\overline{\mathcal{W}}^l$ is used to compute the importance metric. After finishing the compression, we transform the approximated weight $\overline{\mathcal{W}}^l$ to compressed weight $\widetilde{\mathcal{W}}^l$ to the initial weight for the compressed network and then fine-tune the network.
This process is demonstrated in the following figure.

\begin{figure}[h]
\centering
\includegraphics[width=0.9\columnwidth]{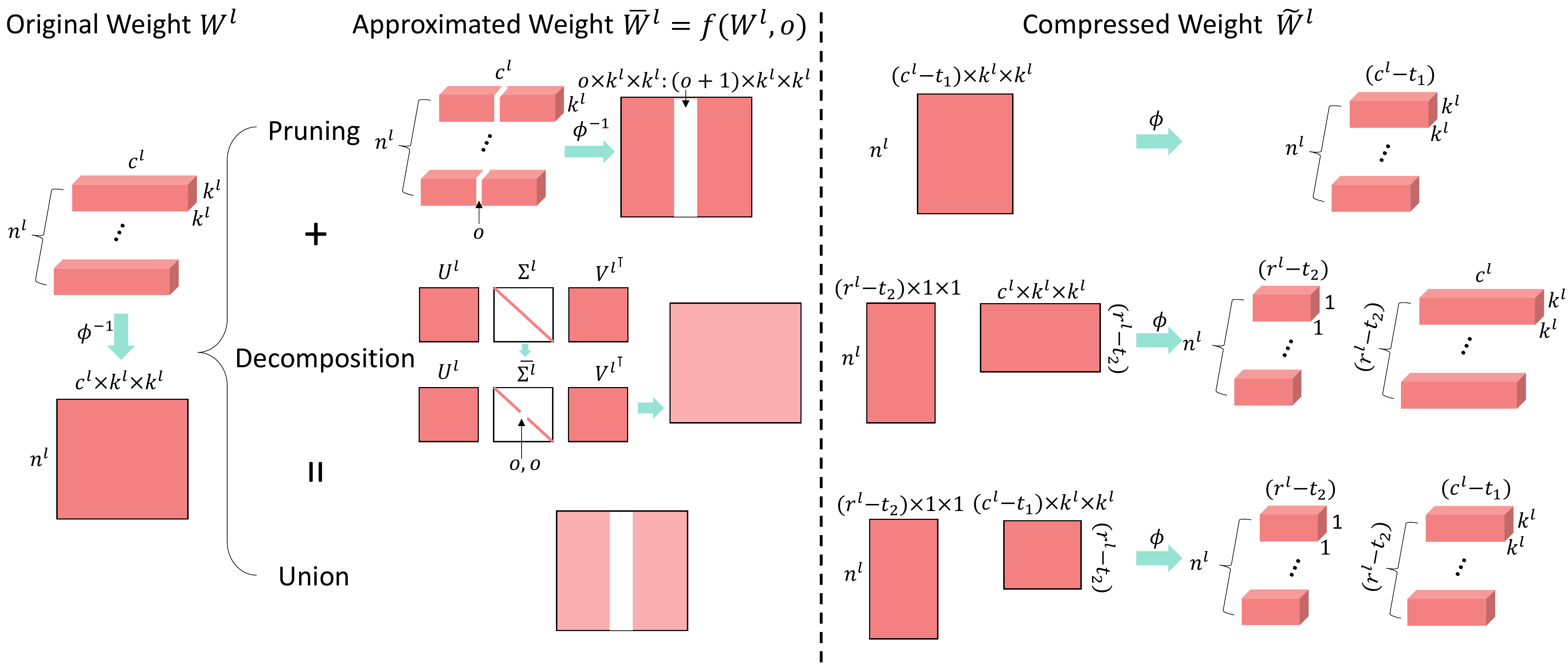}
\caption{Visualization of the compression process.}
      \label{fig:compress}
\end{figure}

\section{Algorithm A}

We provide the heuristic compression algorithm in Algorithm. A.

\begin{algorithm}[H]
 \caption{Heuristic compression algorithm}
 \label{alg:3}
 \begin{algorithmic}[1]
   \REQUIRE{A single layer $\mathcal{W}^l$, average gradient of weight $\mathcal{G}^l$, target compression rate $R^l_a$, calculation interval $T$.}
   \ENSURE{The compressed layer $\overline{\mathcal{W}}^l$.}
   
   \STATE  Initialize the set of compression unit index $\mathcal{U}^l$, whose corresponding unit number is $c^l+r^l$. \\
   \STATE Initialize $\mathcal{W}^{l, 0} = \mathcal{W}^l$, current compression rate $R^l = 0$, current step $t=1$, removed input channels set $\mathcal{U}^{l}_{cp} = \emptyset$. \\
   
    \WHILE{true}
      \FOR {each compression unit index $o$ in $\mathcal{U}^{l,(t)}$}
          \STATE $P^{l,(t)}_{o} = I^{l,(t)}_{o} + \gamma \frac{1}{|\mathcal{U}^{l,(t)}|}\sum\limits_{i \in \mathcal{U}^{l,(t)} \setminus o} I^{l,(t)}_{i|o}$. \\
      \ENDFOR
      \STATE $\overline{\mathcal{W}}^{l,(t)} = \overline{\mathcal{W}}^{l,(t-1)}$. \\
      \STATE $\mathcal{U}^{l,(t+1)} = \mathcal{U}^{l,(t)}$. \\
      \FOR {$T$ least important compression units index $o$ in $\mathcal{U}^{l,(t)}$} 
          \STATE $\overline{\mathcal{W}}^{l,(t)} = f(\overline{\mathcal{W}}^{l,(t)}, o)$. \\
          \STATE $\mathcal{U}^{l,(t+1)} = \mathcal{U}^{l,(t+1)} \setminus o$. \\
          \STATE Compute $R^l$ via Eq.~\ref{eq:compression_ratio_new} in this material.\\
          \IF{$o$ belongs to channel pruning}
            \STATE Add $o$ to $\mathcal{U}^{l}_{cp}$. \\
          \ENDIF
          \IF{$R^l >= R^l_a$}
            \RETURN $\overline{\mathcal{W}}^{l,(t)}$. \\
          \ENDIF
          \IF{$\frac{|\mathcal{U}^{l}_{cp}|}{c^l} >= R^l_a$}
            \STATE $\overline{\mathcal{W}}^l = \mathcal{W}^l$. \\
            \FOR {each compression unit index $o$ in $\mathcal{U}^{l}_{cp}$}
              \STATE $\overline{\mathcal{W}}^l = f(\overline{\mathcal{W}}^l, o)$.\\
            \ENDFOR
            \RETURN $\overline{\mathcal{W}}^l$
          \ENDIF
      \ENDFOR
      \STATE $t = t + 1$. \\
  \ENDWHILE
 \end{algorithmic}
\end{algorithm}

Lines 19-25 in the above algorithm are based on the following analysis.
According to the definition of the compression rate:
\begin{equation}
\small
\label{eq:compression_ratio_new}
R^l = 
\left\{
\begin{aligned}
&1 - \frac{(r^l - t_2)*[(c^l-t_1)*k^l*k^l + n^l]}{n^l*c^l*k^l*k^l} , &t_2 \neq 0; \\
&\frac{t_1}{c^l} , &t_2 = 0,
\end{aligned}
\right.
\end{equation}
if we remove a less number of singular values ($t_2$ is smaller but not equal to zero), the SVD-decomposition will increase the number of parameters, which perhaps leads to extra channel pruning ($t_1$ is larger) to achieve target compression rate.
In contrast, if we only consider channel pruning (\emph{i.e.,} $t_2=0$), $t_1$ will be smaller than the above situation, which keeps more information to achieve the target compression rate.
Therefore, during the compression process, if the weight only compressed by removing input channels has reached the target compression rate (\emph{i.e.,} $\frac{t_1}{c^l}$ larger than the target compression rate), we will only adopt the channel pruning to compress it.

\section{More Comparison with State-of-the-Art Methods}

We compare our method with other methods based on single compression operations for VGG-16 and ResNet-50.
As shown in the following Tab.~\ref{tab:imagenet_new}, compared to GDP \cite{lin2018accelerating}, our method achieves better performance (69.73\% vs. 67.51\%) with higher FLOPs reduction (77.5\% vs. 75.5\%).
Meanwhile, compared to \cite{lin2018accelerating, luo2017thinet, molchanov2019importance, ding2019centripetal, he2019filter, lin2020hrank, lin2019towards}, we also achieve better performance for ResNet-50, which is shown in the following Fig.~\ref{fig:resnet50}.

\begin{table}[h]
\small
\begin{center}
\begin{tabular}{|p{1.5cm}<{\centering}|p{2.5cm}<{\centering}|p{2cm}<{\centering}|p{2.5cm}<{\centering}|p{1.8cm}<{\centering}|p{1.8cm}<{\centering}|}
\hline
Model & Method & FLOPs (PR) & \#Param. (PR) & Top-1 Acc\% & Top-5 Acc\% \\
\hline
\multirow{6}*{VGG-16} & Baseline & 15.48B & 138M & 71.59 & 90.38 \\

 & ThiNet\cite{luo2017thinet} & 9.58B(38.1\%) & 131M(5.1\%) & 69.80 & 89.53 \\

& \textbf{CC($C_t = 0.5$)} & \textbf{7.56B(52.4\%)} & \textbf{131M(5.1\%)} & \textbf{72.05} & \textbf{90.61} \\

 & GDP\cite{lin2018accelerating} & 7.5B(54.5\%) & - & 69.88 & 89.16 \\
 
 & GDP\cite{lin2018accelerating} & 3.8B(75.5\%) & - & 67.51 & 87.95 \\

& \textbf{CC($C_t = 0.75$)} & \textbf{3.48B(77.5\%)} & \textbf{127M(8.0\%)} & \textbf{69.73} & \textbf{89.39} \\

\hline

\end{tabular}
\end{center}
\caption{Comparison with single compression operations-based methods for VGG-16 on ImageNet2012.}
\label{tab:imagenet_new}
\end{table}

\begin{figure}[t]
\centering
\includegraphics[width=0.8\columnwidth]{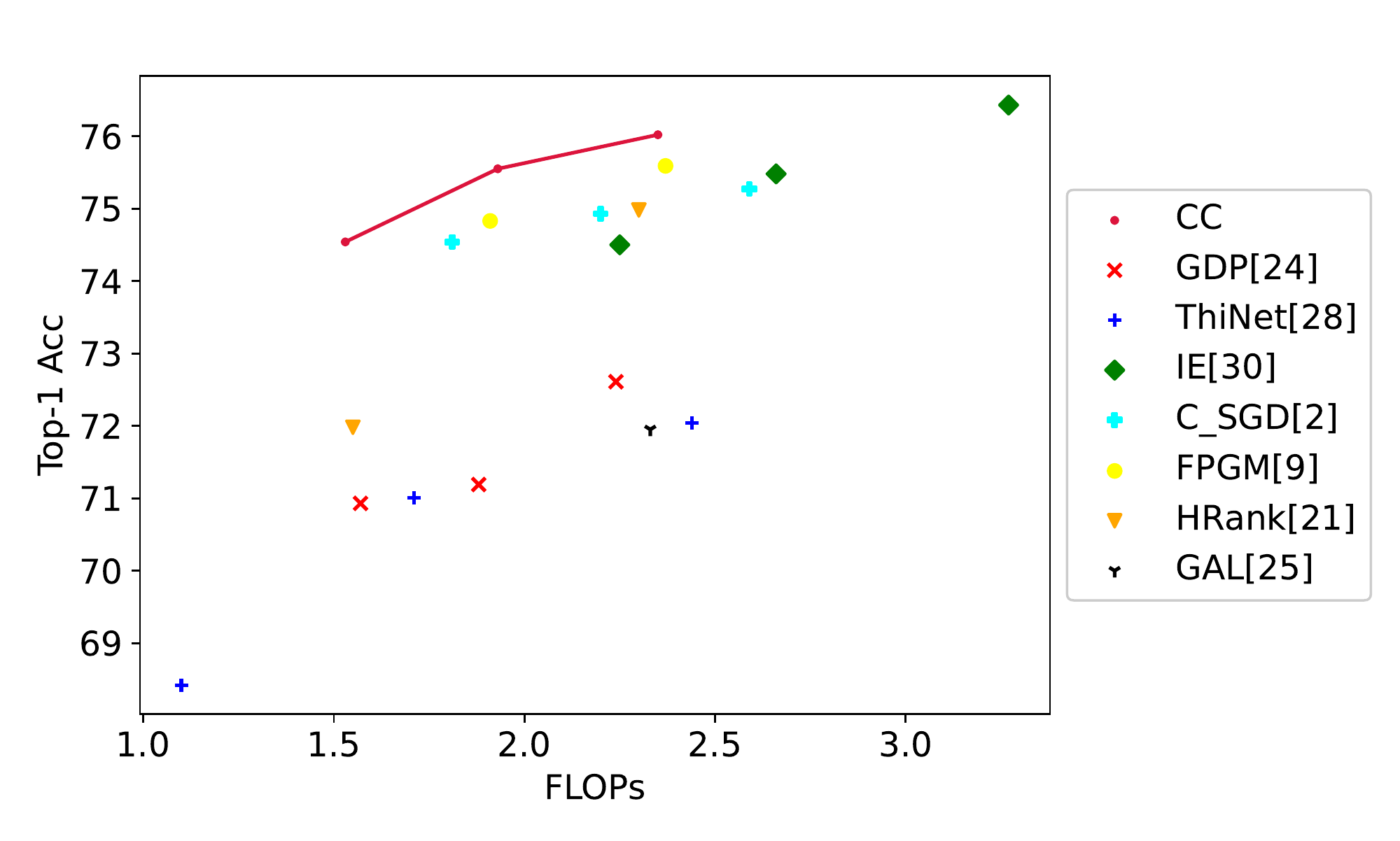}
\caption{Comparison with single compression operations-based methods for ResNet-50 on ImageNet2012.}
      \label{fig:resnet50}
\end{figure}

We evaluate the generalization ability of our method on PASCAL VOC object detection task.
We compress Faster-RCNN with ResNet-50 backbone on Pascal VOC and only obtain 0.85 mAP drop with
50\% compression rate, which demonstrates that our method has a strong generalization ability for the detection task.

\end{document}